%% file: acl_latex.tex
\documentclass[11pt]{article}

\usepackage[final]{acl}

\usepackage{times}
\usepackage{latexsym}
\usepackage{amsmath}
\usepackage{multirow}
\usepackage{makecell}
\usepackage{dcolumn}

\newcolumntype{d}[1]{D{.}{.}{#1}}
\usepackage[T1]{fontenc}

\usepackage[utf8]{inputenc}

\usepackage{microtype}
\usepackage{tabularx}
\usepackage[dvipsnames]{xcolor}
\newcolumntype{Z}{>{\raggedleft\arraybackslash}X}

\usepackage{booktabs}
\usepackage{inconsolata}

\usepackage{graphicx}

\usepackage{lipsum}
\usepackage{amsfonts}
\usepackage{subfig}
\usepackage{todonotes}

\input{tables/DS}

\usepackage{pifont}
\newcommand{\tick}{\textcolor{Black}{\ding{51}}}
\newcommand{\cross}{\textcolor{Black}{\ding{55}}}

\title{Dynamics of Meaning: Towards the Evaluation of Diachronic Semantic Change in Sinhala}

\author{Nevidu Jayatilleke \& Nisansa de Silva \\
  Department of Computer Science \& Engineering, \\
  University of Moratuwa, 
  Sri Lanka \\
  \texttt{\{nevidu.25, NisansaDdS\}@cse.mrt.ac.lk} 
  }

\begin{document}
\maketitle
\begin{abstract}
Tracking semantic change in low-resource languages across extensive historical timelines presents significant challenges due to data scarcity and the limitations of static embedding alignments. This study investigates the diachronic evolution of the Sinhala language from the 13th to the 20th century using a multi-stage computational framework. We first align century-specific \texttt{Word2Vec} and \texttt{FastText} embeddings using Similarity Matrix Based Alignment (SMA) and Orthogonal Procrustes (OP) techniques, finding that OP alignment provides more stable neighbourhood tracking for identifying temporal similarity dips. To move beyond aggregate measures, we introduce a \textit{Bidirectional Semantic Impact Scoring} approach using contextualised embeddings from a fine-tuned \texttt{Llama-3.1-8B}. By applying Leave-One-Out (LOO) diagnostics, we attempt to isolate influential sentences to distinguish between systemic semantic shifts and transient polysemic expansion. 
Our results show that semantic drift in the fine-tuned \texttt{Llama-3.1-8B} is not evenly distributed across all usages. Instead, a significant part of the change is driven by a smaller set of high-impact contextual instances, rather than gradual and uniform change across all occurrences. This work provides a preliminary framework for low-resource Sinhala diachronic analysis, highlighting the trade-offs between model sensitivity and data availability.
\end{abstract}

\section{Introduction}


Language is an evolving system that is continuously reshaped by its speakers. Various factors drive this evolution, including phonetic convenience and sociocultural changes~\cite{keidar2022slangvolution}. In all languages, words naturally exhibit a range of meanings, and the prevalence of these meanings can vary based on the genre and register of discourse, as well as historical context~\cite{frermann-lapata-2016-bayesian}. A well-known example of this is the transformation of the word "gay," which has shifted in meaning from "cheerful" to "homosexual" over the years~\cite{beinborn-choenni-2020-semantic}.



The Sinhala language is an Indo-European language with a rich and diverse literary heritage that has developed over several millennia. Its origins can be traced back to between the 3rd and 2nd centuries BCE. 
The Sinhala language, which is the primary language of the Sinhalese people who constitute the largest ethnic group in the island nation of Sri Lanka, is recognised as the first language (L1) for about 17 million individuals~\cite{de2025survey}. 
Sinhala is classified as a lower-resourced language (Category 02) according to the criteria presented by~\citet{ranathunga-de-silva-2022-languages}. This language has undergone significant evolution and transformation throughout its history, resulting in the modern Sinhala we engage with today.

In this study, we perform a diachronic semantic change analysis using the largest Sinhala diachronic corpus available to date, known as \texttt{SiDiaC-v.2.0}~\cite{jayatilleke2026sidiacv20sinhaladiachroniccorpus}. First, we enhance this corpus to create a lemmatised and POS-tagged version, referred to as \texttt{SiDiaC-v.2.5}. We then develop diachronic embeddings using both static and contextualised methods. This includes various semantic mapping approaches for static embeddings, addressing the issue of extreme data scarcity encountered per century in the corpus. Finally, we provide evaluations of semantic change across centuries based on embedding similarity, while considering the best-performing model-to-mapping combinations for static embeddings and the most effective language model with comprehension capabilities for \texttt{SiDiaC-v.2.0}.

\section{Related Works}

Research into diachronic semantic change has evolved from establishing fundamental statistical laws to developing sophisticated unsupervised detection frameworks. However, there remains a significant gap in these studies concerning data-scarce settings. Specifically, there is a noticeable correlation between corpus size
and the frequency of diachronic semantic change evaluation studies, indicating that studies involving these smaller datasets receive little attention as discussed in Appendix~\ref{app:existing work}.

\subsection{Semantic Drift Evaluation Frameworks}

Foundational work by~\citet{hamilton-etal-2016-diachronic} proposed the "\textit{law of conformity}," which states that frequently used words change slowly, and the "\textit{law of innovation}," which says that polysemous words change rapidly. They also demonstrated that global shifts in word embeddings capture regular linguistic drift, while changes in local neighbourhoods are more effective at detecting irregular cultural shifts~\cite{hamilton-etal-2016-cultural}. 

To overcome the limitations of analysing historical time slices in isolation, \citet{frermann-lapata-2016-bayesian} developed \texttt{SCAN}, a dynamic Bayesian model that monitors the prevalence of word senses over continuous time intervals. By utilising Gaussian Markov Random Fields, the model ensures that adjacent temporal representations are co-dependent, framing semantic shifts as smooth, gradual processes that inherently enhance sparse historical data. Evaluated on a text corpus spanning from 1700 to 2010, this temporal interdependency allowed \texttt{SCAN} to outperform temporally isolated baselines, achieving a Spearman correlation of 0.377 in detecting changes in word meaning, compared to the baseline’s 0.255. Additionally, the model achieved highly competitive results on the \texttt{SemEval-2015} temporal classification benchmarks, reaching an accuracy of 0.748 in 12-year interval predictions.

\citet{rudolph2017dynamic} introduced \textit{Dynamic Bernoulli Embeddings} (\texttt{DBE}) to address the limitations of static word representations, which use a Gaussian random walk prior to model semantic evolution across continuous time slices. By treating embedding vectors as sequential latent variables, this architecture effectively accommodates and smooths out temporally sparse historical data. The dynamic model was evaluated across three diverse corpora, including a 151-year dataset of U.S. Senate speeches that comprises 13.7 million words. It consistently outperformed both static and independently trained time-binned baselines in held-out likelihood predictions. For instance, it achieved a better predictive score of -2.4 compared to the -2.454 score of the time-binned baseline on an \texttt{Arxiv}\footnote{\scriptsize \url{https://arxiv.org/}} dataset. Additionally, by quantifying absolute semantic drift, the model effectively identifies terms that have undergone significant historical changes. Notably, it mathematically identified "Iraq" as the word that experienced the most substantial shift in the Senate corpus, with a peak drift score of 3.09.

For lexical semantics tasks such as \texttt{SemEval-2020}~\cite{schlechtweg-etal-2020-semeval}, systems like \texttt{SenseCluster}~\cite{cuba-gyllensten-etal-2020-sensecluster} utilised multilingual contextualised embeddings (\texttt{XLM-R}) combined with \texttt{K-Means++} clustering. This approach treats clusters as proxies for senses, enabling the identification of both binary and graded changes in meaning. To address the high computational costs associated with deep learning methods and the limitations of analysing only adjacent time periods, recent research has introduced a framework that uses diachronic word similarity matrices derived from aligned \texttt{PPMI-SVD} embeddings~\cite{kiyama-etal-2025-analyzing}. This framework enables a detailed analysis of continuous semantic shifts and facilitates unsupervised clustering of words with similar temporal trajectories.

\subsection{Low-Resource Semantic Drift Evaluation}


Addressing the need for evaluating lexical semantic change in small texts, \citet{ehrenworth-keith-2023-literary-intertextual} explores unsupervised detection models on corpora of about 150,000 tokens. They downsampled the \texttt{SemEval-2020 Task 1}\footnote{\scriptsize\url{https://www.ims.uni-stuttgart.de/en/research/resources/corpora/sem-eval-ulscd/}} datasets while maintaining gold-standard annotations and tested three modelling approaches: static \texttt{SGNS} embeddings with orthogonal Procrustes using cosine distance~\cite{prazak-etal-2020-uwb}, static \texttt{SGNS} with Euclidean distance~\cite{pomsl-lyapin-2020-circe}, and a fine-tuned \texttt{BERT} model with a temporal self-attention mechanism~\cite{rosin-radinsky-2022-temporal}. The results showed a significant 67\% average performance drop in these models in data-scarce environments, though the \texttt{BERT} model remained the most stable. Despite low intrinsic metrics, the researchers highlight the practical utility of these models through a task called "literary intertextual semantic change detection." A case study comparing \texttt{Frantz Fanon}\footnote{\scriptsize\url{https://en.wikipedia.org/wiki/Frantz_Fanon}} and \texttt{Saidiya Hartman}\footnote{\scriptsize\url{https://en.wikipedia.org/wiki/Saidiya_Hartman}} revealed that the contextual model effectively identifies nuanced semantic shifts in terms like "violence" and "political."

To address noise and overfitting in temporally sparse data, \citet{montariol-allauzen-2019-empirical} evaluated three dynamic embedding models (Incremental Skip-Gram (\texttt{ISG})~\cite{mikolov2013distributed, kim-etal-2014-temporal}, Dynamic Filtering of Skip-Gram (\texttt{DSG})~\cite{bamler2017dynamic}, and Dynamic Bernoulli Embeddings (\texttt{DBE})~\cite{rudolph2017dynamic, rudolph2018dynamic} on artificially constrained subsets of the \texttt{New York Times Corpus}~\cite{sandhaus2008new}, simulating data scarcity down to 20,000 words (1\% subset) per year. Their analysis revealed that a "backward external" initialisation, utilising modern, large-scale pre-trained vectors like Wikipedia~\cite{li-etal-2017-investigating} and updating sequentially backward in time, significantly improves performance for sparse historical texts. While the probabilistic dynamic models (\texttt{DSG} and \texttt{DBE}) were far more robust than \texttt{ISG} at capturing directed semantic drifts under extreme scarcity, they initially struggled to isolate extreme shifts. To resolve this, they introduced a \textit{Hardshrink} regularisation term that successfully penalises low-magnitude noise, enabling the models to accurately discriminate valid semantic changes even in severely data-limited environments.

\section{Methodology}

We began by performing lemmatisation and POS tagging on a subset of \texttt{SiDiaC-v.2.0}, specifically focusing on the documents that include written date annotations. For the evaluation of semantic change, we applied two different approaches to this same subset of \texttt{SiDiaC-v.2.0}. The source code of the framework is available on GitHub\footnote{\scriptsize \url{https://github.com/NeviduJ/Semantic-Change-Sinhala}}.

\subsection{Creation of \texttt{SiDiaC-v.2.5}}



Our diachronic analysis utilises the \texttt{SiDiaC-v.2.0}\footnote{\scriptsize \url{https://github.com/NeviduJ/SiDiaC-v.2.0}} corpus~\cite{jayatilleke2026sidiacv20sinhaladiachroniccorpus}, which is the successor to \texttt{SiDiaC-v.1.0}\footnote{\scriptsize \url{https://github.com/NeviduJ/SiDiaC}}~\cite{jayatilleke2025sidiac}. 
\texttt{SiDiaC-v.2.0} contains a total of 229,098 words from 185 documents. 
It is important to note that a book may have been written centuries earlier, while its printed version was released much later~\cite{jayatilleke2025sidiac}. Therefore, for this study, we will focus on a subset of the corpus that includes the written dates, comprising 59 books with a total of 64,813 words, spanning from the 5th century to the 20th century. Further information about these corpora can be found in Appendix~\ref{app:sidiac_corpora}.

\begin{figure}[ht]
    \centering
    \includegraphics[width=0.9\columnwidth]{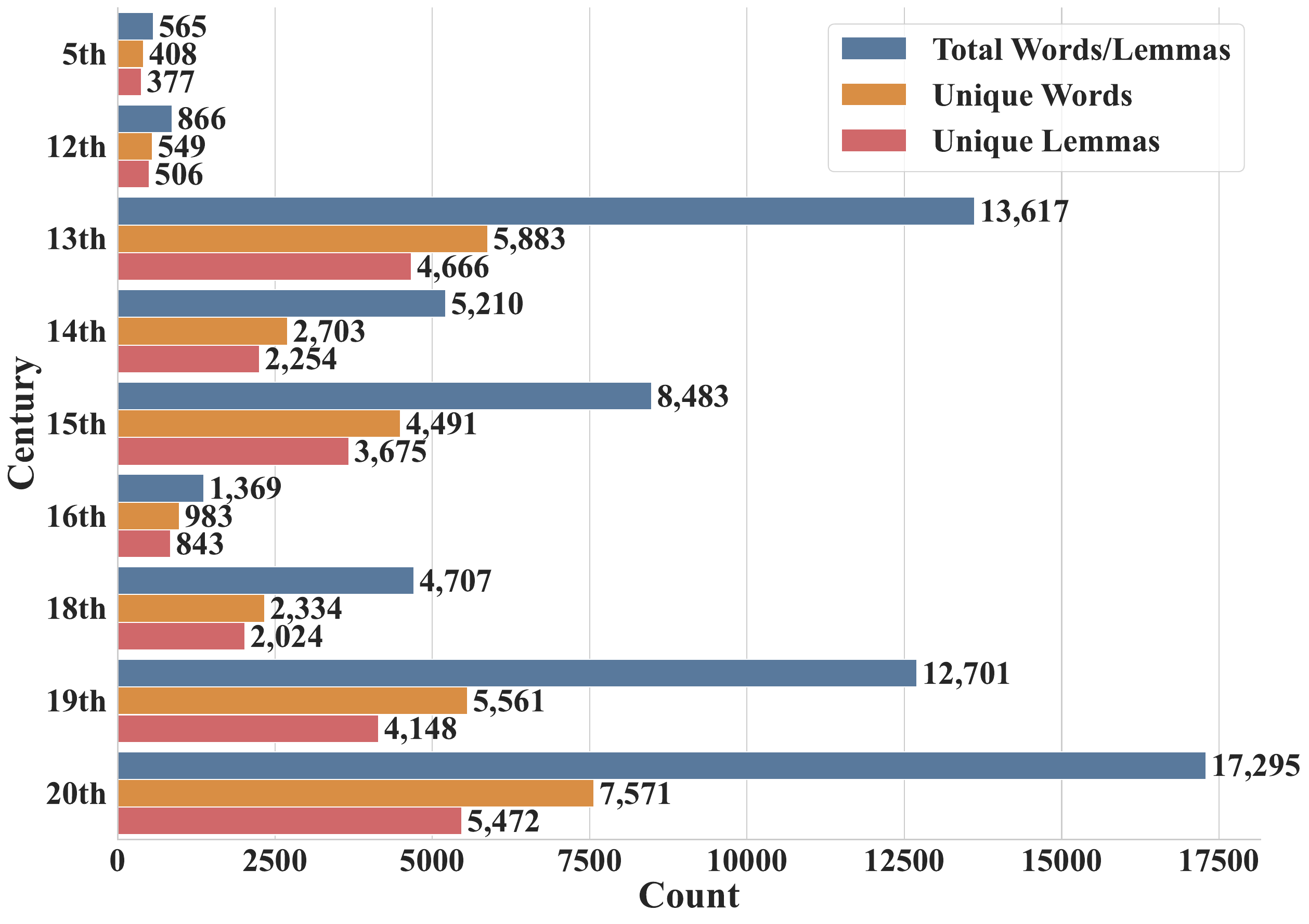}
    \caption{Word Token Counts by Century from the \texttt{SiDiaC-v.2.5} Corpus.}
    \label{fig:tok_count}
\end{figure}


In this study, we enhance the filtered version of the \texttt{SiDiaC-v.2.0} corpus by adding two new features. Specifically, we provide lemmas for each word based on morphological analysis and include POS tags for every word. As a result, we present the improved version of the corpus, an enhanced \textit{Sinhala Diachronic Corpus} (\texttt{SiDiaC-v.2.5})\footnote{\scriptsize \url{https://huggingface.co/datasets/Nevidu/SiDiaC-v.2.5}}.

To create this improved \texttt{SiDiaC-v.2.5}, we utilised \texttt{SinMorphy-v.1.2}\footnote{\scriptsize \url{http://nlp-tools.uom.lk/sin-morphy/}}~\cite{kumarasinghe2021sinmorphy} for lemmatisation. For POS tagging, we developed a composite pipeline that combines \texttt{SinMorphy-v.1.2} with an improved version of the \textit{TnT POS Tagger for Sinhala}\footnote{\scriptsize \url{https://github.com/nlpcuom/Sinhala-POS-Tagger}}~\cite{fernando2018evaluation} and the \texttt{Sinling}\footnote{\scriptsize \url{https://github.com/ysenarath/sinling}} library. A detailed explanation of this enhancement and the new features is provided in Appendix~\ref{app:sidiac_new}.

\subsection{Diachronic Embeddings} 



This section discusses three vectorisation techniques used to construct diachronic word embeddings: \texttt{Word2Vec}, \texttt{FastText}, and transformer-based contextualised embedding models. The 5th, 12th, and 16th centuries were excluded during this semantic change study due to insufficient word counts, as illustrated in Figure~\ref{fig:tok_count}. While the 16th century falls within the 13th--20th century range and was initially considered, its small corpus size reduced the number of consistent words shared across all centuries to 80; removing it raised this to 210. The final analysis, therefore, spans six century-specific corpora: the 13th, 14th, 15th, 18th, 19th, and 20th centuries.

\subsubsection{\texttt{Word2Vec} Embeddings}

\texttt{Word2Vec} embeddings were generated by training a separate model on each century's corpus using a \textit{Skip-gram} architecture~\cite{mikolov2013distributed}, which is better suited to small and sparse corpora as it learns useful representations even for infrequent terms~\cite{hamilton-etal-2016-diachronic}. Each model used a vector size of 50, a context window of 5, and Negative Sampling (\texttt{negative=15}) to handle low-frequency vocabulary. A subsampling threshold of $10^{-5}$ was applied to prevent high-frequency tokens from dominating training~\cite{herbelot-baroni-2017-high}, with the learning rate decaying linearly from $5\times 10^{-2}$ to $7 \times 10^{-4}$ over 200 epochs. This produces distinct word spaces per century, enabling diachronic comparison of word representations across contiguous historical time intervals.

\subsubsection{\texttt{FastText} Embeddings} 

A pre-trained \texttt{FastText} model~\cite{bojanowski2017enriching} trained on the \textit{Sinhala Common Crawl} corpus (\texttt{cc.si.300.bin}) was used as a base model, providing broad lexical coverage grounded in modern Sinhala. The vocabulary of the base model was then extended with any century-specific terms by incremental vocabulary building, and the models were fine-tuned on each century subset for 30 epochs with a linearly decaying learning rate ($1 \times 10^{-2}$ → $1 \times 10^{-4}$). Following training, vocabulary pruning was applied to restrict the model's word vector space to words specifically in that century's corpus, discarding all other words from the pre-trained \texttt{FastText} vocabulary. This procedure yields a set of temporally aligned yet independently fine-tuned embedding spaces. 


\subsubsection{Contextualised Embeddings} 
\label{subsec:context_embd}



We have examined five \textit{Small Language Models} (\texttt{SLMs}), with only one being an encoder-only model (\texttt{XLM-R}) and the remaining four being decoder-only transformers. The decoder-only models include three variants of \texttt{Llama} with 8 billion parameters each (base, instruct, and \texttt{SinLlama}~\cite{aravinda2025sinllama}) and \texttt{Gemma} with 9 billion parameters. We conducted Continual Pre-Training (CPTing) on all models except for \texttt{Gemma}, due to its low performance on perplexity and Bits-Per-Byte (BPB) scores on the non-lemmatised \texttt{SiDiaC-v.2.5} corpus. The CPTing utilised a dataset comprising 90\% training data and 10\% validation data, based on the entire \texttt{SiDiaC-v.2.0} corpus. Additional information on CPTing can be found in Appendix~\ref{app:cpt_tr}.

\begin{table}[h!tb]
\centering
\resizebox{0.9\columnwidth}{!}{
\begin{tabular}{l|c|c|r|c}
\hline
\textbf{Model} & \textbf{Tokeniser} & \textbf{Training} & \textbf{Perplexity} & \textbf{Bits-Per-Byte} \\
\hline
\texttt{Llama-3.1-8B} & \multirow{5}{*}{Base} & \multirow{5}{*}{\cross} & 2.86 & 1.04 \\
\texttt{Gemma-2-9b} & & & 25.17 & 1.49 \\
\texttt{SinLlama} & & & 186.67 & 0.92 \\
\texttt{Llama-3.1-8B-it} & & & 3.42 & 1.22 \\
\texttt{XLM-R} & & & - & 0.80 \\\hline
\texttt{Llama-3.1-8B} & \multirow{3}{*}{Base} & \multirow{3}{*}{\tick} & \textbf{2.18} & \textbf{0.77} \\
\texttt{SinLlama} & & & 2754999.26 & 2.62 \\
\texttt{XLM-R} & & & - & 3.14 \\\hline
\texttt{Llama-3.1-8B} & \multirow{3}{*}{Expanded} & \multirow{3}{*}{\tick} & 623.76 & 1.22 \\
\texttt{SinLlama} & & & 203103.64 & 1.96 \\
\texttt{XLM-R} & & & - & 1.10 \\\hline
\end{tabular}
}

\caption[]{\label{tab:int_eval} Evaluation of SLMs on \texttt{SiDiaC-v.2.0} based on Perplexity and Bits-Per-Byte Metrics.}
\end{table}

Contextual word embeddings were extracted using the best-performing fine-tuned transformer model, the \texttt{Llama-3.1-8B} with base tokeniser, which will be called \texttt{Llama-FT}\footnote{\scriptsize \url{https://huggingface.co/Nevidu/Llama-3.1-8B-SiDiaC-CPT}} from here onwards. This multilingual decoder operates in 4-bit precision and was evaluated based on perplexity and BPB scores presented in Table~\ref{tab:int_eval}. It is important to note that although we expected a better understanding with expanded vocabulary in the tokeniser and then conducting CPTing, it ended up increasing both perplexity and BPB. This may be due to several factors: (1) the addition of more tokens increases the model's surprise, as the next token now has more possible choices; (2) CPTing did not effectively help the model learn the new tokens from the initialised mean position within the latent space, likely due to insufficient data; and (3) the inclusion of whitespace word tokens to expand the vocabulary for models using Byte-Pair Encoding (BPE) for text tokenisation. A detailed analysis of the perplexity and BPB scores can be found in Appendix~\ref{app:int_eval}.

\begin{figure*}[!htbp]
    \centering
        \includegraphics[width=0.77\textwidth]{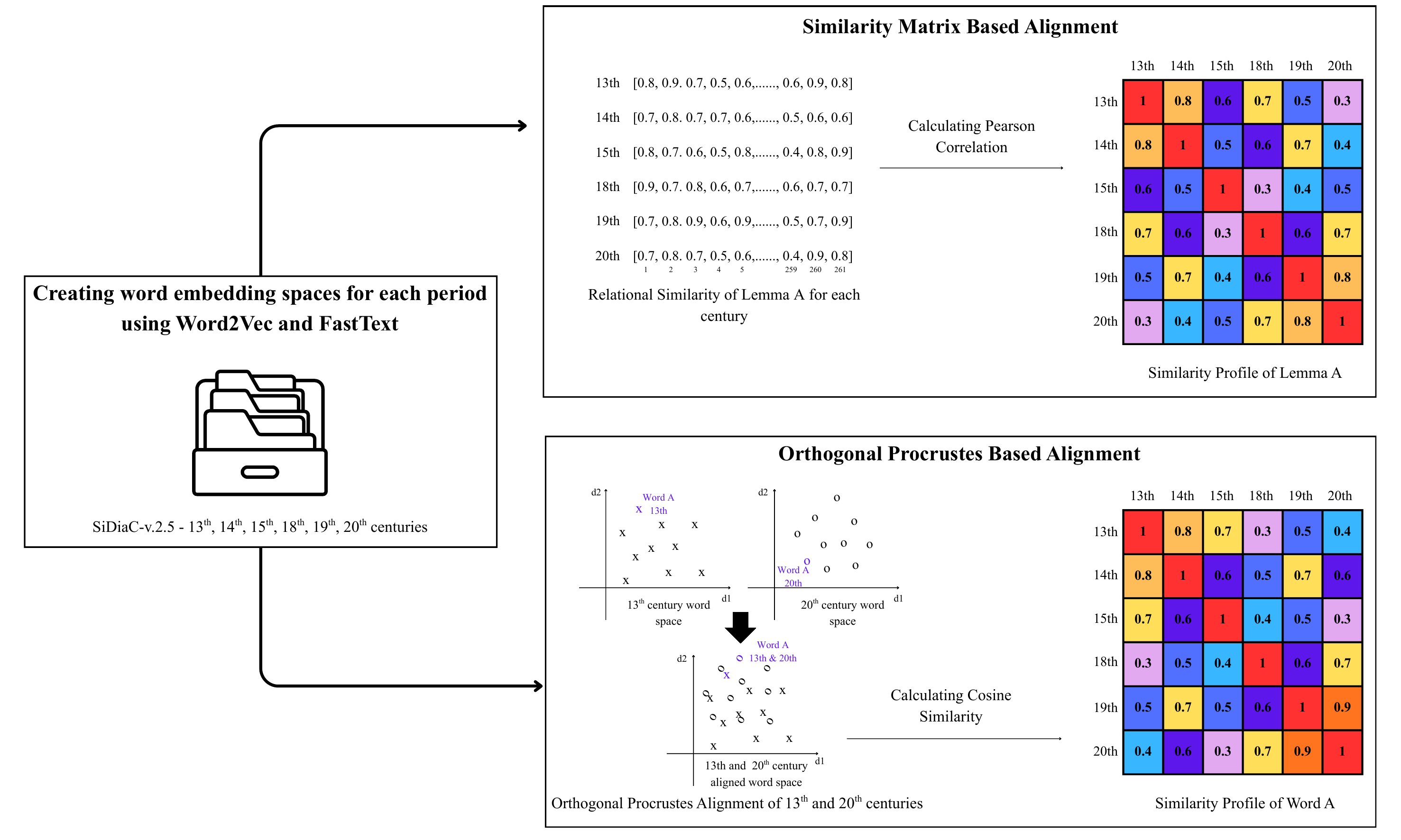}%
    \caption{A framework for analysing diachronic semantic shifts by creating periodic embedding spaces using \texttt{Word2Vec} and \texttt{FastText}, followed by Similarity Matrix-based and Orthogonal Procrustes-based alignment. \\}
    \label{fig:period_embs}
\end{figure*}

The model produces word-level embeddings from the final hidden layer (4096 dimensions). Since \texttt{Llama-FT} employs a subword tokeniser, sub-token vectors for each word were aggregated via mean pooling following the methodology by~\citet{cuba-gyllensten-etal-2020-sensecluster}. To handle long sentences, a sliding window of 510 tokens with a step of 300 tokens was applied, with overlapping regions averaged to reduce boundary effects. This process was applied independently per century for both models, yielding century-specific contextual word representations that capture semantic variation across distinct historical time intervals. It is important to note that, unlike \texttt{Word2Vec} and \texttt{FastText} approaches, where we had trained models for each century, we leverage the frozen pretrained representations following~\citet{martinc-etal-2020-leveraging}. We hypothesise that the high-dimensional embedding space contains inherent low-dimensional manifolds representing century-specific semantic shifts, thereby bypassing the need for period-specific retraining with small datasets, which could lead to overfitting and alignment issues.

\subsection{Computational Semantic Mapping} 
\label{subsec:sem_map}




This section outlines the semantic mapping methodology used to compare word representations across centuries. We employ two complementary alignment strategies for the \texttt{Word2Vec} and \texttt{FastText} embeddings. Such alignment is necessary because these models use randomised initialisation and stochastic training, resulting in vector spaces that are arbitrarily rotated relative to one another~\cite{hamilton-etal-2016-diachronic}, a problem known as non-identifiability~\cite{carrington2019invariance}. A summarised illustration of the alignment mechanisms is shown in Figure~\ref{fig:period_embs}.


\subsubsection{Similarity Matrix Based Alignment}
\label{subsec:sim_matr}

Despite the fact that vector spaces can be ``arbitrarily rotated,'' the similarity between two words within one embedding space can be directly compared to that in another space~\cite{hamilton-etal-2016-diachronic}. This means that even if the axes are rotated, the distance between two vectors remains consistent. Based on this principle, we decided to create a similarity matrix for each word in one space, which can then be compared to another matrix of the same word from a different space.

To account for morphological variation in Sinhala, we first construct a centroid-based lemma representation: for each lemma $L$ with surface forms $\{w_1, w_2, \ldots, w_n\}$, its vector is computed as the arithmetic mean of its valid constituent word vectors, collapsing morphological variance into a single stable point. For cross-century comparison, we adopt a relational similarity approach using 261 common lemmas shared across all century spaces as a common reference frame. Each word is represented by a \textit{similarity profile}, a vector of cosine distances to every anchor lemma, and the \textit{Pearson} correlation between the profiles of the same word in two centuries is used as the measure of semantic similarity, bypassing the need for explicit space alignment. 





\subsubsection{Orthogonal Procrustes Based Alignment}
\label{sec:op_alignment}

The concept of unitary invariance in word embeddings suggests that two embeddings are essentially identical if one can be derived from the other through a unitary operation, such as a rotation~\cite{yin2018dimensionality}. The relative structure and relationships between words remain unchanged if all vectors are transformed by an orthogonal matrix. This preservation of structural integrity ensures that the embeddings retain their usefulness in tasks that rely on cosine similarity, even after these transformations~\cite{carrington2019invariance}. Based on this property, we align century-specific word spaces using two variants of OP analysis: a conventional statistical formulation and a neural network-based formulation.



For rotational alignment, each anchor must be a single, fixed vector in the embedding space, making centroid-based lemma representations unsuitable. The rotation matrix $Q$ relies on a one-to-one correspondence between vectors, so we use the 210 consistent surface-form words as anchors across all century models. This alignment was executed using two approaches:

\paragraph{Analytical Procrustes Alignment:} For each century space, an anchor matrix is constructed from the vectors of these 210 words and subsequently L2-normalised. The optimal rotation matrix $Q = UV^\top$ is derived via Singular Value Decomposition (SVD) of the cross-covariance between the target and reference anchor matrices, rotating the target space to minimise $\|W_{\text{target}} Q - W_{\text{ref}}\|_F$ with respect to the Frobenius-norm while preserving all internal geometric relationships. The 13th-century space serves as the global reference anchor, and all other century spaces are permanently rotated in place to this coordinate system, allowing direct word-level comparison across centuries.


\paragraph{Neural Procrustes Alignment:} As a complementary approach, we frame the OP problem as a supervised learning task using a bias-free, activation-free linear layer $\mathbf{W} \in \mathbb{R}^{d \times d}$, initialised as the identity matrix and trained to map L2-normalised anchor embeddings from a target century into the coordinate system of a reference century. The training objective combines (Mean Squared Error) MSE for alignment with a squared Frobenius-norm orthogonality penalty:
\begin{equation}
    \mathcal{L} = \mathcal{L}_{\text{MSE}} + \lambda \|\mathbf{W}\mathbf{W}^\top - \mathbf{I}\|_F^2 
\end{equation}
where $\lambda = 0.001$ provides the necessary flexibility for the optimiser to minimise alignment error across both \texttt{Word2Vec} and subword-informed \texttt{FastText} architectures without inducing geometric distortion, although higher values of $\lambda$ theoretically enforce a more rigid rotation. The model is optimised with Adam at a learning rate of $0.001$, chosen to ensure stable convergence of the orthogonality constraint. Training runs for up to 2000 epochs with early stopping (patience of 20 epochs, minimum 100 epochs). A separate $\mathbf{W}$ is learned for each target century, with the reference set to the 13th century, yielding five transformation matrices across six centuries.


\begin{table}[h!tb]
\centering
\resizebox{0.9\columnwidth}{!}{
\begin{tabular}{l|c|c|c}
\hline
\textbf{Model} & \textbf{Approach} & \textbf{Avg. Procrustes Disparity} & \textbf{Avg. Rotational Distance} \\
\hline
\multirow{2}{*}{\textbf{\texttt{Word2Vec}}} & Analytical OP & 174.16 & 9.93 \\ \cline{2-4}
& Neural OP & 157.16 & 9.16 \\\hline
\multirow{2}{*}{\textbf{\texttt{FastText}}} & Analytical OP & 57.57 & 22.82 \\ \cline{2-4}
& Neural OP & 57.06 & 19.68 \\ \hline
\end{tabular}
}
\caption[]{\label{tab:eval_proc} Evaluation of Orthogonal Procrustes Alignment Performance.}
\end{table}

We evaluate embedding alignment using \textit{Procrustes Disparity}~\cite{rochkoulets2026entropy}, which measures the residual error $\|W_{target}Q - W_{ref}\|_F^2$ after orthogonal transformation, and \textit{Rotational Distance}, defined as $\|Q - I\|_F$ to quantify the magnitude of the rotation $Q$. As shown in Table~\ref{tab:eval_proc}, the \textit{Neural OP} approach for \texttt{FastText} significantly outperforms the \textit{Analytical OP} used for \texttt{Word2Vec}. This neural-based optimisation achieves better manifold preservation, yielding markedly lower error margins for both disparity and distance metrics compared to the higher absolute residuals of the analytical baseline. While having more data is essential for improving embedding stability, the detailed analysis of the target-to-reference century alignment provided in Appendix~\ref{app:OPA} indicates that \texttt{Word2Vec} worsens with increased data exposure, unlike the fine-tuned \texttt{FastText} models. Therefore, we conclude that \texttt{FastText} should be prioritised in the statistical neighbourhood analysis discussed in \S~\ref{sec:SNA}.





\section{Evaluation of Semantic Change}

\subsection{Statistical Neighbourhood Analysis} 
\label{sec:SNA}

To systematically isolate and qualitatively evaluate lexical items that have undergone semantic change in both the SMA (lemma-level) and the two OP Alignment (word-level) approaches, we implemented an automated reporting mechanism. Due to variations in corpus sizes and differences in the inherent alignment quality across temporal periods, applying a static cosine similarity threshold across all periods was suboptimal. Instead, we established a dynamic, pair-specific approach to define cutoffs for semantic drift.

For any given pair of time periods, we calculate the mean ($\mu$) and standard deviation ($\sigma$) of the cosine similarities for all intersecting vocabulary items between the two aligned vector spaces. A word or a lemma is then classified as having undergone significant semantic change if its similarity score falls below a dynamically defined threshold, calculated as $\mu - (k \times \sigma)$, where $k$ is a scaling factor that controls the stringency of the cutoff. In our analyses, we empirically set $k$ to $0.8$. To validate our choice of $k=0.8$, we examined the proportion of word/lemma instances it flags as candidate drift across all six independent model-to-mapping combinations (\texttt{Word2Vec} and \texttt{FastText}, each under SMA, Analytical OP and Neural OP combinations). Under a Gaussian approximation, a cutoff of $\mu - 0.8\sigma$ is expected to flag the bottom 21.19\% of a distribution ($\Phi(-0.8)$). As shown in Table~\ref{tab:k_threshold_validation} in Appendix~\ref{app:SNA}, the observed flagged proportions ranged from 16.19\% to 22.23\% (mean 18.36\%) across the six combinations, staying within 5 percentage points of the theoretical value in every case despite the combinations differing substantially in embedding model and alignment mechanism. This consistency indicates that $k=0.8$ behaves as a stable, near-canonical proportion of the vocabulary rather than a combination-specific artefact.


\begin{table}[h!tb]
\centering
\resizebox{0.9\columnwidth}{!}{
\begin{tabular}{l|c|c|c}
\hline
 & \textbf{SMA} & \textbf{Analytical OP} & \textbf{Neural OP} \\
\hline
\textbf{\texttt{Word2Vec}} & 30.23\% & 61.31\% & 52.92\% \\
\textbf{\texttt{FastText}} & 55.69\% & 82.11\% & 81.04\% \\ \hline
\end{tabular}
}
\caption[]{\label{tab:mean_sim} Mean similarity of consistent keys across 15 temporal alignments using \texttt{Word2Vec} and \texttt{FastText} for SMA and OP approaches.}
\end{table}

Following the quantitative identification of drifting candidates, we perform a qualitative contextualisation step. For each identified word, we retrieve its top 5 nearest neighbours in both the source and target semantic spaces. By comparing these lists of nearest neighbours side by side, we can manually interpret the specific nature of the semantic shift, observing how a lemma's or a word's contextual usage and associative meanings have transitioned over time. This methodology is applied uniformly across both alignment models, ensuring consistent qualitative insights regardless of the underlying alignment technique.

\begin{figure*}[!htbp]
    \centering
        \includegraphics[width=0.82\textwidth]{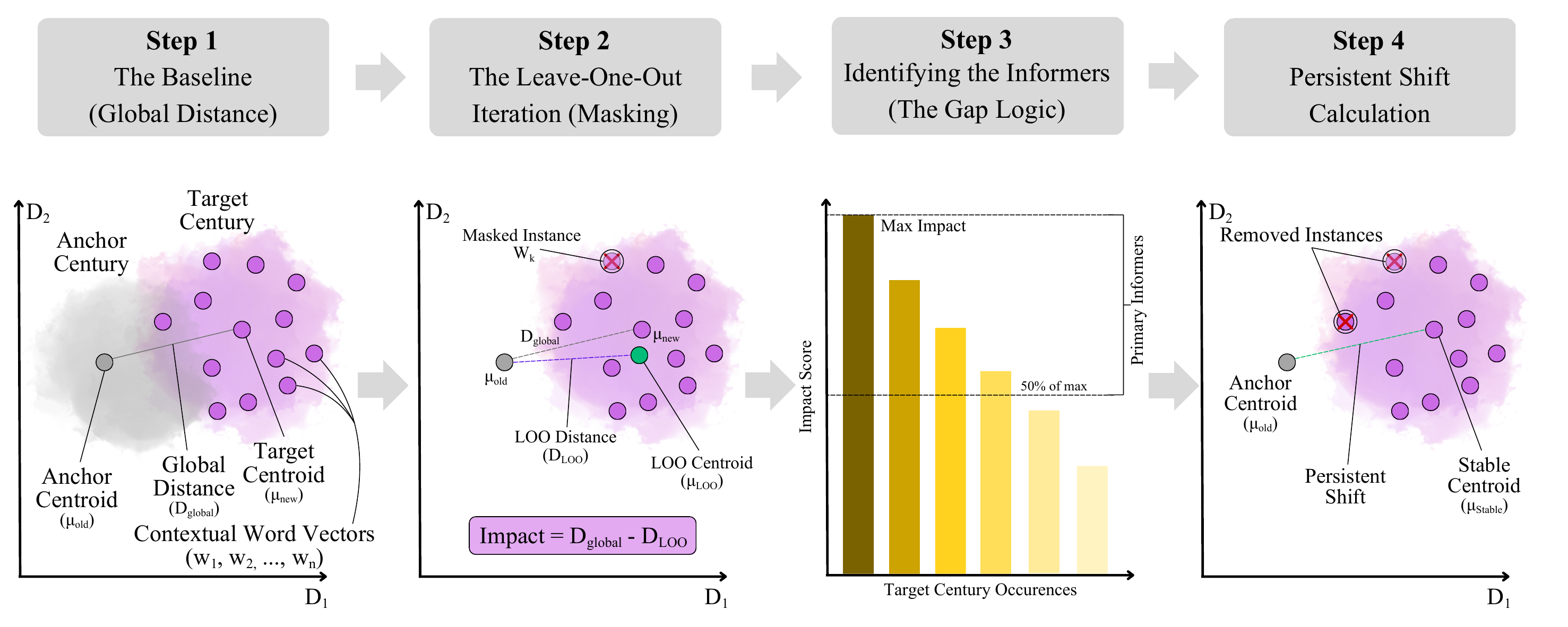}%
    \caption{Leave-One-Out (LOO) analysis for identifying contextual semantic divergence. \\ \small \textbf{Step 1:} The baseline global distance is established. \textbf{Step 2:} Individual contextual word vectors ($w_k$) are iteratively masked to measure their impact on centroid displacement. \textbf{Step 3:} Textual evidence of semantic divergence is extracted as ``primary informers'' based on this impact. \textbf{Step 4:} Informers are permanently removed to isolate the word's persistent shift in core meaning.}
    \label{fig:LOO}
\end{figure*}

We evaluate all six combinations using \texttt{Word2Vec} and \texttt{FastText} across three alignment strategies: SMA and two OP variants. Following \citet{gulordava-baroni-2011-distributional}, we expect high mean similarity in temporal semantic spaces, as most of the lexicon remains stable while semantic change affects only a small vocabulary subset. Our results indicate high similarity in \texttt{FastText} methods as shown in Table~\ref{tab:mean_sim}. However, this could be due to fine-tuning on pretrained models to compensate for the small corpus. Thus, we must consider both \texttt{Word2Vec} and \texttt{FastText}, given their unique strengths and weaknesses.



In contrast, SMA appears less robust based on the mean similarity statistics in Table~\ref{tab:mean_sim} compared to OP techniques. This is likely because SMA considers all consistent lemmas when calculating relational similarity. This global approach includes both stable anchor lemmas and those that have undergone semantic change, meaning the resulting relational similarity shifts are distorted by those localised drifts.

We analysed a total of 15 century-aligned semantic spaces, each containing 261 consistent lemmas and 210 consistent words. This resulted in 2,116 instances of similarity for all combinations of models and alignment approaches for each century-to-century alignment. During our analysis, we found that 229 instances of these lemmas were captured using the \texttt{Word2Vec} model and 59 using \texttt{FastText} through the SMA approach. Additionally, we identified 635 word instances from \texttt{Word2Vec} and 271 from \texttt{FastText} using the analytical procrustes approach. Furthermore, the neural procrustes approach revealed 648 word instances from \texttt{Word2Vec} and 274 from \texttt{FastText}.

The high frequencies of temporal similarity dips observed in the \texttt{Word2Vec} embeddings, along with the similarly distributed frequencies across the 15 aligned semantic spaces as shown in Table~\ref{tab:dip_analysis}, again indicate that \texttt{Word2Vec} embeddings were less stable compared to \texttt{FastText} embeddings.

For example, using \texttt{FastText} and neural procrustes alignment, we can observe that the word ‘\raisebox{-0.5ex}{%
\includegraphics[height=1.5\fontcharht\font`\A]{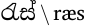}}’ in the 14th century appears in a semantic environment associated with gathering, movement, and central positioning. Its neighbors include causing to gather (\raisebox{-0.5ex}{%
\includegraphics[height=1.4\fontcharht\font`\A]{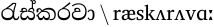}}), receiving or accepting (\raisebox{-0.5ex}{%
\includegraphics[height=1.3\fontcharht\font`\A]{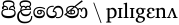}}), in the middle (\raisebox{-0.5ex}{%
\includegraphics[height=1.3\fontcharht\font`\A]{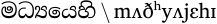}}), and with speed (\raisebox{-0.5ex}{%
\includegraphics[height=1.5\fontcharht\font`\A]{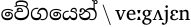}}). In this context, `\raisebox{-0.5ex}{%
\includegraphics[height=1.5\fontcharht\font`\A]{Figures/si_161.pdf}}' is linked with the idea of collecting or assembling, often in physical or spatial terms. By the 15th century, the lemma shifts into a more symbolic and doctrinal semantic field. Its neighbors include destruction or end (\raisebox{-0.5ex}{%
\includegraphics[height=1.4\fontcharht\font`\A]{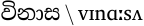}}), life span (\raisebox{-0.5ex}{%
\includegraphics[height=1.3\fontcharht\font`\A]{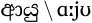}}), secret or hidden (\raisebox{-0.5ex}{%
\includegraphics[height=1.4\fontcharht\font`\A]{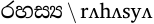}}), gem or precious object (\raisebox{-0.5ex}{%
\includegraphics[height=1.5\fontcharht\font`\A]{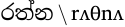}}), and the Buddha’s radiance (\raisebox{-0.5ex}{%
\includegraphics[height=1.5\fontcharht\font`\A]{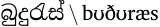}}). In this period, `\raisebox{-0.5ex}{%
\includegraphics[height=1.5\fontcharht\font`\A]{Figures/si_161.pdf}}' is also associated with spiritual and symbolic meanings, especially light, brilliance, and sacred radiance in Buddhist contexts. Further examples from all six model-to-mapping combinations are discussed in Appendix~\ref{app:SNA} as part of a small-scale manual validation.

\subsection{Bidirectional Semantic Impact Scoring}
\label{sec:loo}

By using created \texttt{Llama-FT} contextual embeddings, filtered to the six valid centuries between 13th--20th as discussed in \S~\ref{subsec:context_embd}, lemma-level analysis was enabled by constructing a sentence-aligned lemma lookup cross-referencing the lemma and word token versions of the corpus: for each occurrence of a word, identified by its sentence and position within each century, we mapped the corresponding lemma to its contextual word vectors and the original sentence texts. This process created a flat record for each lemma occurrence that includes its century, surface-form words, vectors, and original sentence contexts.

From this, a lemma map was constructed per century: for each lemma, all contextual vectors of its surface form words across all sentences were collected, preserving the full distributional variation of each lemma within a century. The lemma centroid, computed as the arithmetic mean of all instance vectors, serves as the aggregate representation for cross-century comparison. Critically, the individual instance vectors and their source texts are retained alongside the centroid.

To interpret the specific linguistic context driving the numerical change observed between chronological centroids generated from \texttt{Llama-FT} embeddings, we implemented a diagnostic Leave-One-Out (LOO) analysis. The baseline for semantic change is established as the global cosine distance between the anchor century's centroid and the target century's centroid ($D_{global}$). To isolate the impact of individual textual usages, the algorithm iterates through every vectorised instance of the lemma in the target century. In each iteration, a single sentence instance is provisionally excluded from the target pool, a new centroid ($D_{LOO}$) is calculated, and the cosine distance to the anchor centroid is re-evaluated.

By subtracting $D_{LOO}$ from the baseline global distance ($D_{global}$), we quantify the semantic impact of the excluded sentence. A positive impact score indicates that the presence of the sentence actively pulled the target centroid away from its historical anchor meaning. To capture related senses of semantic change, the algorithm identifies the sentence with the maximum displacement impact and flags it, alongside any other instances yielding an impact score within 50\% of this maximum, as primary ``informers''. The 50\% threshold was chosen as a mid-level salience criterion to balance specificity and coverage when identifying important contextual usages. Higher thresholds often selected only one dominant sentence, while lower thresholds included many weakly influential instances, which reduced interpretability. Therefore, the 50\% cutoff keeps instances whose impact is still close to the strongest effect, capturing the main contributors to semantic change without adding noise. In a forward temporal comparison (lower to upper century), these ``informers'' represent novel semantic innovations; in a reverse comparison (upper to lower century), they highlight obsolete contextual usages that eventually faded from the lexicon.

Finally, out of the total instance count for the target lemma, these isolated informer sentences are permanently masked from the target century representations to recalculate a revised distance metric, termed the ``persistent shift''. The mathematical difference between the original global distance and the persistent shift determines exactly what percentage (Drift Reduction) of the observed semantic drift is attributable to specific documented contextual innovations. A summary of the complete mechanism is illustrated in Figure~\ref{fig:LOO}. This LOO methodology, applied uniformly across \texttt{Llama-FT} embeddings over all 30 bidirectional permutations of the six centuries, provides both quantitative measurement and immediate qualitative textual evidence of semantic change.

The results show a low but consistent level of semantic shift in the \texttt{Llama-FT} representations. The average global centroid distance is 0.0295, which decreases to 0.0272 after removing high-impact instances, resulting in a drift reduction of 8\%. This indicates that a measurable portion of the observed shift is due to a subset of high-influence contextual usages. Out of 112,945 instances, a total of 21,738 are identified as having drifted, revealing that semantic change is unevenly distributed across occurrences. 

As an example, the semantic trajectory of the lemma "\raisebox{-0.5ex}{%
\includegraphics[height=1.4\fontcharht\font`\A]{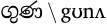}}" (guna, meaning "virtue/quality") reveals a clear divergence between the 14th-century context and the 15th-century context. Both periods share a core of "virtue-praise," but they differ in whose virtue is celebrated and in what manner. In the 14th century, "\raisebox{-0.5ex}{%
\includegraphics[height=1.4\fontcharht\font`\A]{Figures/si_201.pdf}}" is used in a style that praises kings, princes, and wise people, describing them as figures of high moral character and many virtues. It also refers to a great sage known for having all virtues. This usage is supported by related terms such as clan/lineage (\raisebox{-0.5ex}{%
\includegraphics[height=1.3\fontcharht\font`\A]{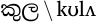}}), city (\raisebox{-0.5ex}{%
\includegraphics[height=1.5\fontcharht\font`\A]{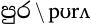}}), compassion (\raisebox{-0.5ex}{%
\includegraphics[height=1.4\fontcharht\font`\A]{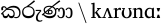}}), and endowment (\raisebox{-0.5ex}{%
\includegraphics[height=1.5\fontcharht\font`\A]{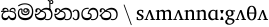}}, as in "endowed with twelve/ten virtues"), reflecting a blend of dynastic and religious praise vocabulary. In the 15th-century context, the usage of "\raisebox{-0.5ex}{%
\includegraphics[height=1.4\fontcharht\font`\A]{Figures/si_201.pdf}}" broadens along two distinct lines: the continued and intensified Buddhist compassion-praise (e.g., praising the Buddha's abundant loving-kindness) and courtly-poetic praise of feminine beauty, which describes various heroines, including a princess named Ulakudaya Devi and a brief reference to Sita (\raisebox{-0.5ex}{%
\includegraphics[height=1.4\fontcharht\font`\A]{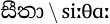}}). This poetry often compares gentle virtues to the pure light of the moon. The linguistic landscape of this period shifts to include terms related to fame/renown (\raisebox{-0.5ex}{%
\includegraphics[height=1.5\fontcharht\font`\A]{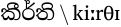}}) and triumph (\raisebox{-0.5ex}{%
\includegraphics[height=1.5\fontcharht\font`\A]{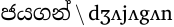}}), as seen in a passage describing a heroine's beauty as "triumphing over" the celestial wish-tree. Meanwhile, compassion (\raisebox{-0.5ex}{%
\includegraphics[height=1.4\fontcharht\font`\A]{Figures/si_204.pdf}}) remains a common thread connecting both centuries. The eleven example sentences that underlie the 27.31\% drift reduction reflect a similar mix of themes, Buddhist compassion-praise alongside praise for feminine beauty rather than a singular, clear narrative. Both strands pull the focus away from the 14th century's masculine, royal-heroic sense. This pattern consistently recurs across four separate century pairs for this lemma. Instead of a complete shift in meaning, this represents the closest form of semantic change provided by the corpus: a stable abstract sense of "virtue/quality," whose objects of praise expand from masculine, royal-heroic figures to include both intensified religious devotion and feminine courtly beauty.

Overall, these findings support the idea that diachronic semantic drift in this model is driven by selective contextual innovations rather than uniform shifts across all instances. A detailed analysis of these metrics, based on all 30 bidirectional permutations of the six centuries, along with two more examples of semantic change, is discussed in Appendix~\ref{app:BSIS} as part of a small-scale manual validation.


\section{Conclusion}

This study established a computational framework for diachronic semantic analysis using three distinct embedding methodologies. We successfully aligned century-specific \texttt{Word2Vec} and \texttt{FastText} models through SMA and OP alignment techniques. Our evaluation of these static embeddings via statistical neighbourhood analysis demonstrated the superior stability of \texttt{FastText} followed by OP alignment for identifying temporal ``similarity dips'' in the Sinhala lexicon.

Furthermore, we extended the analysis beyond aggregate centroid-based measures by introducing \textit{Bidirectional Semantic Impact Scoring}. Utilising contextualised embeddings from \texttt{Llama-FT}, this approach leveraged LOO diagnostics to isolate and identify sentences responsible for semantic drift. In contrast to global averaging, this method attempts to distinguish between systemic shifts and polysemic expansion by highlighting specific influential instances. While limited by available data density, the approach offers a preliminary framework for exploring the semantic evolution of low-resource languages like Sinhala across historical periods.

\section*{Limitations}

The scope and outcomes of this study on Sinhala diachronic semantics were influenced by several constraints.

\paragraph{Data Scarcity:} The lack of extensive historical corpora impacted the strength of century-wise \texttt{Word2Vec} models, leading to possible problems such as unreliable \textit{Procrustes} alignment compared to the fine-tuned \texttt{FastText} models. Additionally, the transformer models did not improve the perplexity and BPB scores during continual pretraining, possibly due to data scarcity affecting the embedding strength of expanded vocabulary words.

\paragraph{Lemmatisation:} \texttt{SinMorphy} may not have been trained on historical Sinhala. The absence of robust morphological analysers highlights the low-resource status of Sinhala~\cite{de2025survey}.

\paragraph{POS Tagging:} The absence of high-accuracy POS taggers for historical Sinhala~\cite{de2025survey} necessitated a hybrid tagging system. This system likely carries a modern linguistic bias, as its dependencies are rooted in contemporary Sinhala grammar, potentially affecting the accuracy of morphological labelling within the medieval and early modern sections of the corpus.

\paragraph{WiC (Word-in-Context) Models:} This LLM approach, commonly used in diachronic semantic drift research, is designed to mimic the behaviour of human annotators when assessing the similarity of word meanings in pairs of sentences from two different contexts, C1 and C2. The embeddings $\Phi_1$ and $\Phi_2$ are extracted from the trained WiC model for computational comparisons~\cite{periti2024lexical}. While it could be argued that these language models, which are based on pretrained models like \texttt{XLM-R} that support Sinhala, can be used in their raw form, fine-tuning on WiC datasets that lack Sinhala coverage may lead to catastrophic forgetting of language-specific knowledge. This concern is highlighted by~\citet{cassotti-etal-2023-xl}, that the WiC model \texttt{XL-LEXEME} tends to perform poorly on languages that are underrepresented in the training set of \texttt{XLM-R} and not included in the WiC dataset used during training. Therefore, we chose not to use these models in our study.

\section*{Acknowledgment}

This research study was made possible through the valuable contributions of several individuals and organisations. We would like to express our gratitude to the \textit{LK Domain Registry} for the funding provided to publish this paper. We sincerely thank \texttt{Vihindi Kotalawa} for her support in conducting the extended literature review of diachronic corpora. Additionally, we appreciate \texttt{Dr. Surangika Ranathunga} and \texttt{Dr. Aloka Fernando} for sharing the repository for the \textit{TnT POS Tagger for Sinhala}, which was essential for developing the proposed POS tagging pipeline.

\bibliography{anthology-1, anthology-2, custom, lrec2026-example}

\appendix

\section{Corpora vs Semantic Drift Studies}
\label{app:existing work}

\begin{figure*}[!htbp]
    \centering
    
    \begin{minipage}{0.48\textwidth}
        \centering
        \subfloat[81 listed diachronic corpora.]{
            \includegraphics[width=\textwidth]{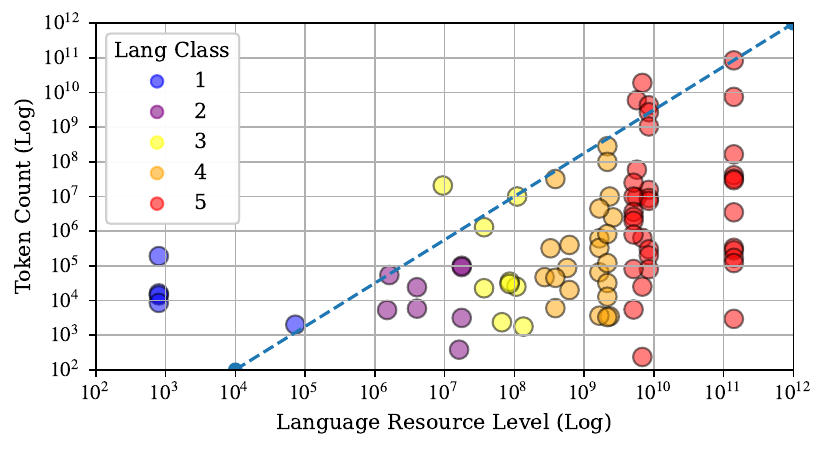}
            \label{fig:Copora_all}
        }
    \end{minipage}
    \hfill 
    \begin{minipage}{0.48\textwidth}
        \centering
        \subfloat[Corpora that have undergone semantic drift evaluation.]{
            \includegraphics[width=\textwidth]{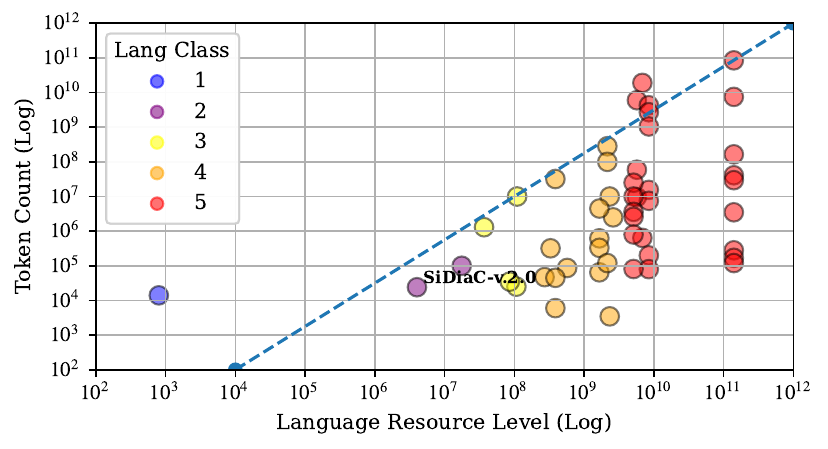}
            \label{fig:Copora_filtered}
        }
    \end{minipage}
    
    \caption{Log token counts of the existing diachronic corpora listed by \citet{jayatilleke2026sidiacv20sinhaladiachroniccorpus} (including \texttt{SiDiaC-v.2.0}) against the log of language resource level.}
    \label{fig:Copora_master}
\end{figure*}

\begin{figure*}[!htbp]
    \centering
    
    \begin{minipage}{0.31\textwidth}
        \centering
        \subfloat[Undergone statistical semantic drift evaluation.]{
            \includegraphics[width=\textwidth]{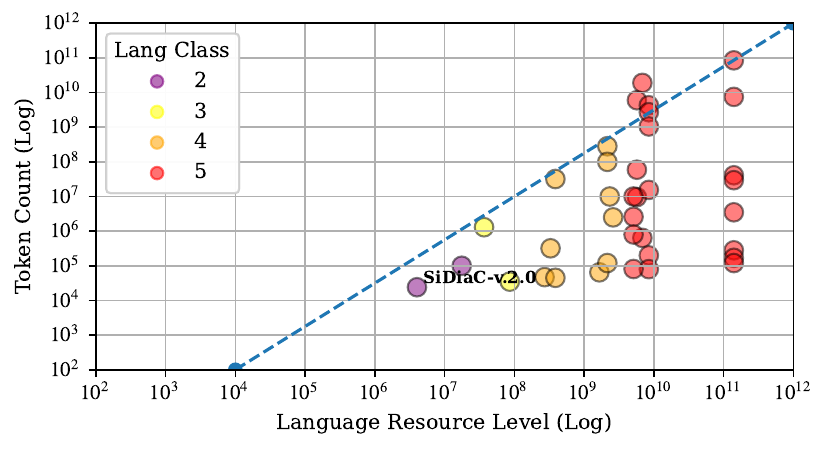}
            \label{fig:Copora_Stat}
        }
    \end{minipage}%
    \hfill 
    \begin{minipage}{0.31\textwidth}
        \centering
        \subfloat[Undergone qualitative semantic drift evaluation.]{
            \includegraphics[width=\textwidth]{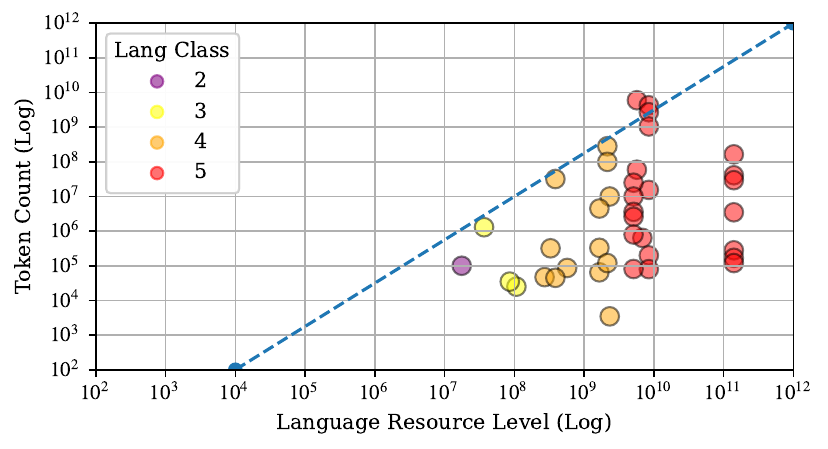}
            \label{fig:Copora_Qual}
        }
    \end{minipage}%
    \hfill
    \begin{minipage}{0.31\textwidth}
        \centering
        \subfloat[Undergone embedding based semantic drift evaluation.]{
            \includegraphics[width=\textwidth]{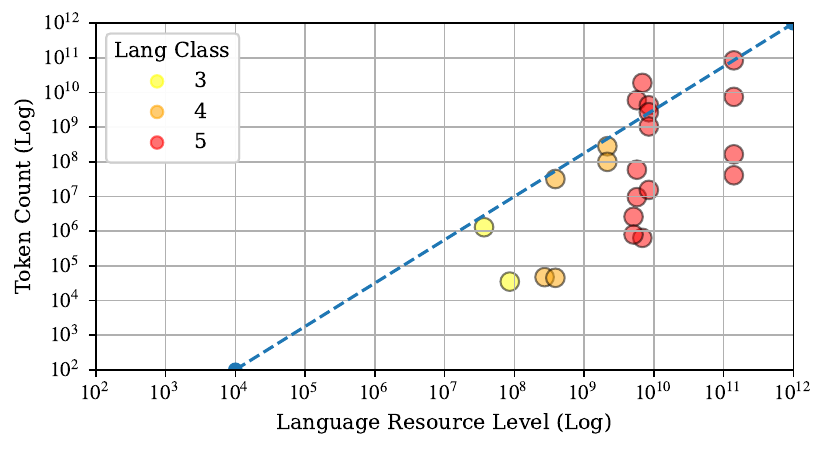}
            \label{fig:Copora_Embed}
        }
    \end{minipage}
    
    \caption{Log token counts of the existing diachronic corpora listed by \citet{jayatilleke2026sidiacv20sinhaladiachroniccorpus} (including \texttt{SiDiaC-v.2.0}) against the log of language resource level - classified based on evaluation approach.}
    \label{fig:Corpora_eval_master}
\end{figure*}


In this study, we analysed the corpora used for diachronic semantic change research based on the list identified by \citet{jayatilleke2026sidiacv20sinhaladiachroniccorpus}. A total of 81 corpora were listed, including the proposed \texttt{SiDiaC-v.2.0}. The phenomena examined in the works surveyed by \citet{tahmasebi2018survey} regarding computational approaches to lexical semantic change detection are tagged under the following headings: lexical (semantic) change, grammaticalisation, and lexical replacement. Based on these tags, we identified 50 corpora that have been used at least once in a semantic change study.

Based on our findings, it is important to highlight that \texttt{SiDiaC-v.2.0} is one of three corpora that have undergone studies on semantic change for any language classified below language class 3, according to the language resource taxonomy by \citet{ranathunga-de-silva-2022-languages} as shown in Figure~\ref{fig:Copora_master}. Among these three corpora, \texttt{SiDiaC-v.2.0} is unique in its general study of lexical semantic change, as it does not focus on specific instances of grammaticalisation like the other two corpora. \citet{jayatilleke2026sidiacv20sinhaladiachroniccorpus} conducted this semantic change evaluation on \texttt{SiDiaC-v.2.0} using a statistical Bag of Words (BoW) analysis during the proposal of the corpus. This emphasises that low-resource languages are often overlooked in the field of diachronic semantic change, highlighting a significant gap in this area of research. 

Furthermore, it is noteworthy that out of the 50 corpora that have undergone evaluations of semantic drift, 37 have employed statistical methods while 39 have involved qualitative assessments that draw on human expertise. Interestingly, only 22 have used distributional representations (embeddings) for evaluation, and no corpora below language resource class 3 have been evaluated using this method, making our study the first to address this gap for a low-resource class 2 language. This information is detailed in Figure~\ref{fig:Corpora_eval_master}. 

A comprehensive summary of the surveyed corpora by~\citet{jayatilleke2026sidiacv20sinhaladiachroniccorpus}, along with their status regarding lexical semantic change studies and several example research articles, can be found in Table~\ref{tab:AllDS}.

\begin{table*}[h!tb]
    \centering
    \resizebox{0.95\textwidth}{!}{
        \begin{tabular}{|l|l|r|c|c|c|c|c|c|c|}
        \hline
        \multirow{2}{*}{\textbf{Dataset}} & \multicolumn{2}{c|}{\textbf{Language}} & \multirow{2}{*}{\textbf{Time Span}} & \multirow{2}{*}{\textbf{Token Count}} & \multirow{2}{*}{\textbf{LSC Status}} & \multicolumn{3}{c|}{\textbf{LSC Evaluation Approach}} & \multirow{2}{*}{\textbf{Example LSC Studies}}\\
        \cline{2-3}\cline{7-9}
         & \textbf{Name} & \textbf{Class} &  &  &  & \textbf{Statistical} & \textbf{Qualitative} & \textbf{Embedding} &\\
        \hline
        \DSTable
        \end{tabular}
        }
    \caption{Summary of the \LitCount{} diachronic corpora surveyed by~\citet{jayatilleke2026sidiacv20sinhaladiachroniccorpus}; including and sorted by the language class as defined by~\citet{ranathunga-de-silva-2022-languages}.}
    \label{tab:AllDS}
\end{table*}

\section{\texttt{SiDiaC}: Sinhala Diachronic Corpus}
\label{app:sidiac_corpora}

\subsection{\texttt{SiDiaC-v.1.0}}

\texttt{SiDiaC-v.1.0}~\cite{jayatilleke2025sidiac} is the first comprehensive diachronic corpus for the Sinhala language, consisting of approximately 58,000 word tokens extracted from 46 literary works that span from the 5th to the 20th century CE. The dataset was created by digitising texts acquired from the National Library of Sri Lanka using the \texttt{Google Document AI}~\cite{jayatilleke-de-silva-2025-zero} OCR engine. This engine performed crucial functions such as text modernisation and morpheme segmentation to effectively manage historical orthography. 

A key feature of \texttt{SiDiaC-v.1.0} is its annotation based on the written date, which was determined through the lifespans of authors and secondary historical sources like the \textit{Sinhala Sahithya Wanshaya}~\cite{Sannasgala_2009}, rather than relying on the printed issue date. This approach ensures a precise timeline for historical analysis, despite the challenges associated with dating undated manuscripts.

The corpus is organised with a two-level genre classification system: a primary layer that distinguishes between Fiction and Non-Fiction, and a secondary layer that categorises texts into Religious, History, Poetry, Language, and Medical genres. The content reflects Sri Lanka's literary history, showcasing a predominance of religious and poetic texts due to the influence of Theravada Buddhism and courtly patronage. It also retains code-mixed passages in Pali and Sanskrit to preserve the original context. 

\subsection{\texttt{SiDiaC-v.2.0}}

\texttt{SiDiaC-v.2.0}~\cite{jayatilleke2026sidiacv20sinhaladiachroniccorpus} is the largest comprehensive diachronic corpus for Sinhala to date, significantly expanding upon its predecessor. It includes 229,098 word tokens across 185 literary works. The full corpus spans publication dates from 1800 to 1955 CE; however, a critically important subset of 59 documents, comprising 64,805 words (though our study recorded 64,813 words, likely due to minor variations in automated preprocessing), has been annotated with precise written dates ranging from the 5th to the 20th century CE. This annotation facilitates accurate historical analysis.

Similar to \texttt{SiDiaC-v.1.0}~\cite{jayatilleke2025sidiac}, text extraction was performed using the \texttt{Google Document AI}~\cite{jayatilleke-de-silva-2025-zero} OCR engine. This tool was chosen for its advanced capabilities in text modernisation and morpheme segmentation, which are necessary for processing historical Sinhala orthography. The corpus follows a two-level genre classification system. It first distinguishes between Fiction and Non-Fiction and then further categorises texts into specific domains such as Religious, History, Poetry, Language, and Medical. A frequency analysis illustrating the distribution of resources across genres and written time periods, represented in centuries, is shown in Table~\ref{tab:dist_cent_sidiac}.

\begin{table*}[h!tb]
\centering
\resizebox{0.9\textwidth}{!}{
\begin{tabular}{r|cc|ccccc|r}
\hline
& \multicolumn{2}{c|}{\textbf{Primary Category}} 
& \multicolumn{5}{c|}{\textbf{Secondary Category}} &
\multirow{2}{*}{     \textbf{Total}}\\
\cline{2-8}
& \multicolumn{1}{c}{\textbf{Non-Fiction}}& \multicolumn{1}{c|}{\textbf{Fiction}} & \multicolumn{1}{c}{\textbf{Religious}} & \multicolumn{1}{c}{\textbf{Poetry}} & \multicolumn{1}{c}{\textbf{Language}} & \multicolumn{1}{c}{\textbf{History}} & \multicolumn{1}{c|}{\textbf{Medical}} & \\
\hline
\textbf{5th} & 1 & 0 & 0 & 0 & 0 & 0 & 1 & 1 \\
\textbf{12th} & 0 & 1 & 0 & 1 & 0 & 0 & 0 & 1 \\
\textbf{13th} & 12 & 1 & 8 & 1 & 3 & 1 & 0 & 13 \\
\textbf{14th} & 2 & 2 & 3 & 0 & 0 & 1 & 0 & 4 \\
\textbf{15th} & 4 & 4 & 3 & 3 & 2 & 0 & 0 & 8 \\
\textbf{16th} & 1 & 0 & 0 & 1 & 0 & 0 & 0 & 1 \\
\textbf{18th} & 4 & 0 & 2 & 0 & 1 & 0 & 1 & 4 \\
\textbf{19th} & 7 & 3 & 4 & 3 & 2 & 1 & 0 & 10 \\
\textbf{20th} & 12 & 5 & 6 & 6 & 2 & 2 & 0 & *16 \\
\hline
\textbf{Total} & 43 & 16 & 26 & 15 & 10 & 5 & 2 & 59 \\
\hline
\end{tabular}}
\caption[]{\label{tab:dist_cent_sidiac}
Distribution of Books by Written Centuries and Genres in \texttt{SiDiaC-v.2.5}.\\
{\small *The total count for the secondary category in the 20th century is 16, while the overall number of books is 17. This discrepancy occurs because the book `\textit{Hithopadhesha Sannaya}', which offers advice, was not classified under any of the five secondary categories. It is important to note that this distributional analysis is based on \texttt{SiDiaC-v.2.0}~\cite{jayatilleke2026sidiacv20sinhaladiachroniccorpus}, which serves as the foundation for \texttt{SiDiaC-v.2.5}.}}
\end{table*}


To ensure linguistic integrity, the dataset underwent deeper post-processing by native speakers. This process involved removing code-mixed content (including Pali, Sanskrit, and English) and correcting formatting issues. Additionally, special tokens such as \texttt{<eos>} for sentence boundaries and \texttt{<psi>} to preserve the unique suffix structure of Sinhala poetry were introduced.

\section{Creation of \texttt{SiDiaC-v.2.5}}
\label{app:sidiac_new}

\subsection{Lemmatisation}
\label{app:lematisation}

Lemmatisation was performed on the filtered \texttt{SiDiaC-v.2.0} subset using \texttt{SinMorphy-v.1.2}\footnote{\scriptsize \url{http://nlp-tools.uom.lk/sin-morphy/}}~\cite{kumarasinghe2021sinmorphy}, which is a morphological analyser. Since \texttt{SinMorphy} is accessible only through a web interface, we used that platform to perform morphological analysis on our sentences from each century. The automation process operates using a \texttt{Selenium}\footnote{\scriptsize \url{https://www.selenium.dev/}} based browser instance to sequentially process tokenised sentences. It incorporates adaptive retry logic to reduce the impact of network latency and potential timeouts during server interaction. 

\texttt{SinMorphy}~\cite{kumarasinghe2021sinmorphy} is a rule-based system that relies on a \textit{Finite State Transducer} (\texttt{FST})~\cite{beesley2003finite} to process words independently, without employing deep learning or contextual information. The system is implemented using the \texttt{Foma}\footnote{\scriptsize \url{https://fomafst.github.io/}}~\cite{hulden-2009-foma} compiler for general rules and the \texttt{LEXC} file format to define the lexicon and specify morphological continuation classes. Its internal structure comprises four main components: a lexicon categorised by POS, a set of morphemes, lexical tags, and rules that dictate operations such as concatenation, deletion, and gemination. For words not explicitly listed in the vocabulary, the system employs a ``guesser'' component that inspects the final characters of the unknown word to infer and assign likely POS tags based on known morphological patterns.


For each word, we parse the morphological string to isolate the root lemma, selecting the appropriate analysis when multiple analyses are provided. We select all options if the system is uncertain about the POS tag, which is disambiguated after the final POS tagging discussed in \S~\ref{subsec:pos_tagging}. The system also addresses irregular outputs, particularly for unrecognised terms, which are primarily labelled as ``proper noun or invalid'' by \texttt{SinMorphy}. In these cases, it defaults to using the original word form when no explicit lemma is provided. To further ensure data integrity, we conducted a filtration step that scans each extracted lemma for validity, where any lemma that contains non-Sinhala characters is flagged as erroneous and replaced with the original word. 



\subsection{POS Tagging}
\label{subsec:pos_tagging}

During lemmatisation with \texttt{SinMorphy}, the tool assigned probable POS tags based on established morphological patterns, as detailed in Appendix~\ref{app:lematisation}. However, there were instances where it assigned multiple possible POS tags. For example, the word `\raisebox{-0.5ex}{%
    \includegraphics[height=1.45\fontcharht\font`\A]{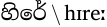}%
}' was tagged as either a noun or a verb. To determine the correct POS tags in these cases, we utilised the \textit{TnT POS Tagger for Sinhala}\footnote{\scriptsize \url{https://github.com/nlpcuom/Sinhala-POS-Tagger}}~\cite{fernando2018evaluation}.

\textit{Trigrams`n'Tags} (\texttt{TnT})~\cite{brants-2000-tnt} is a statistical tool that employs the \textit{Viterbi algorithm}~\cite{forney2005viterbi} to implement second-order Markov models for POS tagging. It generates taggers based on \textit{Hidden Markov Models} (\texttt{HMMs}) that calculate the most likely sequence of grammatical tags for a given text. To improve the accuracy of these probability estimates and effectively address data sparsity, the method uses linear interpolation as a smoothing technique.

The \texttt{TnT} (\texttt{HMM}) model was trained on two manually annotated datasets: a \textit{News corpus} containing 200,000 words from various sections of local newspapers (such as general news, foreign news, and sports) and an \textit{Official Document corpus} consisting of 83,000 words from government circulars, reports, and letters. Both corpora were annotated using the second level of the ``\textit{Comprehensive Multilevel Sinhala POS Tag Set}~\cite{fernando2016comprehensive},'' which features 31 distinct tags designed to accommodate the vocabulary differences between the semi-spoken style of news and the formal written style of official documents.

During the implementation of the \textit{TnT POS Tagger for Sinhala}, we had to make several adjustments to meet our requirements, where we improved the system to make it more robust and efficient. First, we faced the unavailability of the original C-based \texttt{TnT} tagger. To address this, we emulated the \texttt{TnT} behaviour using the \texttt{nltk.tag.tnt}\footnote{\scriptsize \url{https://www.nltk.org/api/nltk.tag.tnt.html}} module. Secondly, we encountered an efficiency issue with the standard \textit{Viterbi algorithm}~\cite{forney2005viterbi} used in the model. This algorithm explores every possible path through the \texttt{HMM} states, resulting in a complexity of approximately $O(T^L)$ for a sentence of length $L$ and tagset size $T$. To improve this, we subclassed `\texttt{nltk.tag.tnt.TnT}' and overrode the `\texttt{tag()}' method with an iterative \textit{Beam Search algorithm}. In this mechanism, at each step for a word, we rank all possible partial paths based on their log-probability and retain only the top $N$ best paths, determined by the beam width. We set the beam width to 100, which significantly reduces complexity to linear $O(L \cdot N \cdot T)$. This adjustment allows for instantaneous tagging of long sentences without a significant loss in accuracy.

Due to the presence of textual content that spans across different time periods, the lexicon lacked certain words in the corpus, resulting in 23,180 words being labelled as `\textit{UNK}'. To address this issue, we utilised another POS tagger from the \texttt{Sinling}\footnote{\scriptsize \url{https://github.com/ysenarath/sinling}} library to specifically identify the POS tags of the words that had been assigned multiple tags by \texttt{SinMorphy}.

To address ambiguities in POS tagging and morphological analysis, we employ a priority-based selection algorithm that integrates multiple data sources. For each token, we extract potential POS tags along with their corresponding morphological structures from \texttt{SinMorphy}. In cases where multiple POS candidates are given, the algorithm disambiguates the token by first consulting the \textit{TnT} tagger, with the \textit{Sinling} tagger serving as a fallback option. 

The disambiguation process follows a strict hierarchy to ensure consistency between high-level POS categories and low-level morphological features. Initially, raw POS tags from secondary taggers such as \texttt{TnT} and \texttt{Sinling} are mapped to a standardised set of seven main-level categories, including \textit{NOUN}, \textit{VERB}, \textit{PART}, \textit{PROPER NOUN}, \textit{ADJECTIVE}, \textit{PRONOUN}, and \textit{FOREIGN WORD}, while unidentified symbols are categorised as UNKNOWN. For every token, the morphological analyser provides a list of possible tags and a corresponding list of morphologies, which are maintained in parallel to ensure each tag remains linked to its correct linguistic interpretation. 

The decision logic dictates that if only one candidate is available, it is automatically selected. In cases of ambiguity where multiple candidates exist, the algorithm searches for a match between the mapped \texttt{TnT} result and the candidate set. If the \texttt{TnT} result is unknown or does not match any candidate, the algorithm repeats this search using the mapped \texttt{Sinling} result. If no valid match is found from either tagger, the system defaults to the first candidate in the analyser's output to maintain data integrity. Once the final tag is determined, its specific index is used to retrieve the matching morphological string. The process concludes with lemma extraction, where the string is split at the first occurrence of the ``+'' delimiter to isolate the leading root substring.

\section{Contextualised Embeddings}
\label{app:cont_emb}

\subsection{Continual Pre-training (CPTing)}
\label{app:cpt_tr}

The applied CPTing strategies were optimised for the distinct structural mechanisms of causal and masked language models. To evaluate the downstream impact of structural vocabulary adaptation across disparate model families, we selected \texttt{Llama-3.1-8B} and its downstream Sinhala variant \texttt{SinLlama} as our Causal Language Model (CLM) representative, and \texttt{XLM-R} as our Masked Language Model (MLM) representative. For both architectures, we conducted dynamic vocabulary expansion by extracting out-of-vocabulary (OOV) terms from the entire \texttt{SiDiaC-v.2.0} corpus via an optimised set-difference lookup and also conducted CPTing with the native base tokenisers of the respective models. However, due to structural divergence in how autoregressive models construct representations versus how bidirectional encoders capture context, the subsequent optimisation, initialisation, and training strategies are decoupled into architecture-specific tracks.

For the CLM stream (\texttt{Llama-3.1}, and \texttt{SinLlama}), the approach focuses on structural stabilisation to prevent representation collapse within the inner transformer blocks. We integrated experimental combinations with both base tokenisers and newly added tokens (expanded tokenisers) into the network by initialising their weights using the exact average vector computed from the model's existing embedding matrix. For experiments involving expanded tokenisers, we implemented a dedicated embedding warmup phase lasting for a single epoch. During this phase, all core attention layers were locked, and Low-Rank Adaptation (LoRA) matrices were applied exclusively to the input token embeddings and the final language modelling heads to help align the new coordinates into a stable semantic space. Once we achieved boundary stability, these parameters were mathematically merged back into the base layers. Next, we initiated deep contextual learning optimised for memory, using 4-bit quantisation (this stage was the same for both native base tokeniser-based and expanded tokeniser-based experiments). A comprehensive LoRA configuration was then deployed across all attention projection layers and Multi-Layer Perceptron (MLP) layers over three epochs, utilising early stopping with a patience of three, monitoring validation loss every 50 steps.

Conversely, the MLM stream (\texttt{XLM-R}) is adapted using an end-to-end, full-parameter optimisation loop to align the expanded token structures with its pre-existing multilingual representations. Unlike the autoregressive track, XLM-R's native architecture lacks parameter wrappers, enabling direct structural surgery on its standard embedding layers. The model is expanded to its final terminal vocabulary size (totalling $307,386$ tokens after adding $57,384$ tokens) and trained directly on an MLM objective where a standard 15\% token masking probability is enforced. Because this process tunes the model's bidirectional contextual weights globally, all encoder layers are unfrozen and optimised simultaneously. 

\subsection{Intrinsic Evaluation}
\label{app:int_eval}

To evaluate the linguistic processing efficiency of the models on \texttt{SiDiaC-v.2.5} corpus, we calculate Perplexity (PPL) and Bits Per Byte (BPB) using a sliding window approach optimised for long sequences. For each sequence, we compute the total Negative Log-Likelihood (NLL) by shifting an evaluation window of size $W$ by a scoring stride $S$. To isolate the evaluation to the target stride and prevent context bleeding from inflating the prediction quality, tokens belonging to the historical context window are explicitly masked by setting their target labels to $-100$. Because standard deep learning frameworks compute the cross-entropy loss as an average over active targets, we reconstruct the true segment NLL by scaling the model's average loss output by the precise count of non-masked target tokens within that window, denoted as $N_{\text{active}} = \sum \mathbb{I}(\text{target}_j \neq -100)$. Accumulating these values over the entire evaluation corpus, the final corpus-level metrics are computed as:

\begin{equation}
\text{PPL} = \exp\left( \frac{\sum_{k} \text{NLL}_k}{\sum_{k} N_{\text{active}, k}} \right)
\end{equation}

\begin{equation}
\text{BPB} = \frac{\sum_{k} \text{NLL}_k}{B_{\text{total}} \cdot \ln(2)}
\end{equation}

where $\text{NLL}_k$ and $N_{\text{active}, k}$ represent the absolute negative log-likelihood and active target count for the $k$-th sliding window segment, respectively, and $B_{\text{total}}$ is the absolute byte length of the raw, UTF-8 encoded text corpus. While perplexity remains inherently dependent on the specific tokenisation vocabulary size due to its per-token normalisation, the derivation of \texttt{BPB} yields a highly reliable, tokeniser-agnostic benchmark that enables direct cross-architectural evaluation.


The impact of tokenisation on perplexity scores is clearly illustrated in Table~\ref{tab:int_eval} when comparing the results of fine-tuned models versus base models. In the conducted CPTing experiments, where we first expanded the vocabulary, we observed a significant disparity of perplexity scores compared to the standard versions of the corresponding models. A larger vocabulary provides a greater range of options for selecting the next token, which spreads out the model's probability distributions. As a result, the model faces more difficulty in confidently choosing the correct option, leading to increased uncertainty and higher perplexity scores. Unfortunately, after conducting CPTing, the BPB scores also increased, though not as significantly as the perplexity scores. This indicates a need for more data for this type of task, even after utilising the entirety of \texttt{SiDiaC-v.2.0}, which comprises approximately 229k words.

Among standard decoder-only architectures, \texttt{Llama-3.1-8B} exhibits superior baseline performance across both metrics. However, when evaluating \texttt{SinLlama}, a model that has been continuously pre-trained on a 10-million-token Sinhala corpus using an expanded tokeniser, it's clear that the architecture of the tokeniser significantly impacts both perplexity and BPB evaluations. Although BPB is inherently tokeniser-agnostic because it is normalised over raw byte counts, its value is highly sensitive to the structural inefficiencies introduced by vocabulary expansion. Specifically, expanding a vocabulary to encompass entire words or specialised sub-words leads to an excessively sparse output probability space. This substantial softmax overhead dilutes the model's confidence in its predictions, resulting in increased total cross-entropy loss. Consequently, when normalised, the model displays a higher BPB, revealing the explicit information-theoretic penalty of compressing the same raw text with an oversized vocabulary. This structural penalty illustrates why CPTing using the native base tokeniser yields the best scores for both perplexity and BPB in the default \texttt{Llama-3.1-8B} configuration. Maintaining a compact, original vocabulary preserves prediction density and minimises entropy, even when processing fragmented low-resource scripts. 

Following the approach by~\citet{wang-etal-2024-improving-text} for using instruct models to generate embeddings, we experimented with the instruct variant of \texttt{Llama-3.1-8B}. Unfortunately, this led to lower scores in both perplexity and BPB compared to the base model.

In our analysis, the only encoder-only model we considered, \texttt{XLM-R}, achieved the second-best BPB score. It is important to note that \texttt{XLM-R} does not produce traditional perplexity scores. This is because its bidirectional masked language modelling objective computes loss only for isolated masked tokens, rather than modelling the joint probability of an entire sequence of text. This information highlights that, despite the emergence of larger models, an older encoder-only model is still powerful at grasping the nuances of low-resource languages. Based on both BPB and perplexity scores, we concluded that \texttt{Llama-FT} is the model to use for the bidirectional semantic impact scoring evaluation discussed in \S~\ref{sec:loo}.

\section{Computational Semantic Mapping}
\label{app:csm}

Two semantic mapping approaches for \texttt{Word2Vec} and \texttt{FastText} were utilised for century-wise semantic analysis, as discussed in \S~\ref{subsec:sem_map}. The alignment mechanisms are summarised and illustrated in Figure~\ref{fig:period_embs}.

\subsection{Orthogonal Procrustes Alignment}
\label{app:OPA}

The OP alignment was conducted using two distinct approaches: analytical and neural, as described in \S~\ref{sec:op_alignment}. We evaluated the alignment approaches based on two metrics: Procrustes disparity and rotational distance. In this section, we will examine the metric results obtained for each model-to-mapping combination corresponding to the alignment from the reference century to the target century. This analysis clarifies the average scores presented in Table~\ref{tab:eval_proc}.

\begin{table*}[h!tb]
\centering
\resizebox{0.8\textwidth}{!}{
\begin{tabular}{c|c|c|c|c}
\hline
\textbf{Model} & \textbf{Approach} & \textbf{Target Century} & \textbf{Procrustes Disparity} & \textbf{Rotational Distance}\\
\hline
\multirow{10}{*}{\texttt{Word2Vec}} 
 & \multirow{5}{*}{Analytical OP} 
   & 14th & 172.37 & 9.93 \\
 & & 15th & 173.04 & 10.09 \\
 & & 18th & 168.95 & 9.83 \\
 & & 19th & 177.85 & 9.82 \\
 & & 20th & 178.58 & 9.98 \\
\cline{2-5}
 & \multirow{5}{*}{Neural OP} 
   & 14th & 153.12 & 9.14 \\
 & & 15th & 159.41 & 9.30 \\
 & & 18th & 140.86 & 8.77 \\
 & & 19th & 165.56 & 9.22 \\
 & & 20th & 166.86 & 9.35 \\
\hline
\multirow{10}{*}{\texttt{FastText}} 
 & \multirow{5}{*}{Analytical OP} 
   & 14th & 59.73 & 22.75 \\
 & & 15th & 56.81 & 22.82 \\
 & & 18th & 69.72 & 22.63 \\
 & & 19th & 51.91 & 22.97 \\
 & & 20th & 49.68 & 22.95 \\
\cline{2-5}
 & \multirow{5}{*}{Neural OP} 
   & 14th & 59.15 & 19.71 \\
 & & 15th & 56.33 & 19.79 \\
 & & 18th & 68.38 & 19.47 \\
 & & 19th & 51.80 & 19.72 \\
 & & 20th & 49.62 & 19.72 \\
\hline
\end{tabular}}
\vspace{0.4em}
\caption{\label{tab:procrustes_all} Evaluation of Analytical and Neural OP alignment performance of \texttt{Word2Vec} and \texttt{FastText} embedding spaces, reported per target century.\\ \scriptsize \textbf{Note:} Each target century's embedding space is aligned to the 13th-century embedding space, which serves as the reference for all alignments.}
\end{table*}

Our analysis reveals a fundamental theoretical conflict in how data density affects vector space stability across the two static embedding architectures, as detailed across Figure~\ref{fig:tok_count} and Table~\ref{tab:procrustes_all}. Standard machine learning theory dictates that increased training data should improve structural stability and alignment precision. This expected behaviour is observed clearly in the fine-tuned \texttt{FastText} models. As data density increases from the 18th century (4,707 tokens) to the 20th century (17,295 tokens), the Procrustes Disparity following analytical OP drops significantly from its peak error of 68.38 down to its global minimum of 49.62 in \texttt{FastText}. 

In contrast, \texttt{Word2Vec} exhibits the exact opposite trend, providing results that appear theoretically incorrect. When data volume increases from the 18th to the 20th century, \texttt{Word2Vec}'s metrics degrade simultaneously: its Procrustes Disparity rises from 140.86 to its absolute maximum of 166.86, while its Rotational Distance increases from 8.77 to a peak of 9.35 following neural OP alignment, as shown in Table~\ref{tab:procrustes_all}. A similar trend is noted in \texttt{Word2Vec} following analytical OP alignment regarding Procrustes Disparity, though the Rotational Distance appears somewhat scattered, showing no clear trend. This means that instead of converging toward the 13th-century baseline with more text data, the \texttt{Word2Vec} space shifts further away and requires more global coordinate rotation to achieve an alignment compromise. In a from-scratch word-level model, scaling up the token count introduces a severe vocabulary expansion problem. Rather than simply refining existing word paths, extra data introduces thousands of completely unique, rare word context windows. This forces the high-dimensional space to expand and distort dynamically to fit these new parameters. Consequently, when forced back onto the 13th-century anchor via a rigid linear transformation matrix $Q$, the true semantic drift and vocabulary expansion accumulated over 7 centuries are fully exposed, yielding a higher disparity and requiring greater rotational force. 

Conversely, \texttt{FastText} avoids this distortion because it is fine-tuned. Its underlying subword manifolds are bound by prior geometric constraints that dictate where words must reside. Additional tokens simply resolve and smooth out the pre-existing coordinate space rather than chaotically expanding it. This enables a gradual reduction in disparity as more data is added, while maintaining a stable rotational distance at around 19.68 across all centuries, as shown in Table \ref{tab:procrustes_all} following neural OP alignment. Similar trends can be observed following analytical OP where rotational distance is at around 22.82. Thus, while \texttt{Word2Vec} fails to provide theoretically good alignment metrics, its rising scores are an accurate, unconstrained mathematical footprint of the actual structural evolution of a language learned from scratch.

The higher Rotational Distance in \texttt{FastText} ($\approx 19.72$) compared to \texttt{Word2Vec} ($\approx 9.35$) following neural OP alignment is caused by how the models are initialised. Because \texttt{Word2Vec} is learned from scratch, its coordinate system is highly flexible. It easily rotates locally to match small vocabularies with very little structural effort. 

In contrast, \texttt{FastText} inherits a rigid, pre-trained subword structure. This pre-existing manifold binds the entire language space together as a single block, preventing it from collapsing or shifting easily. Aligning this large, pre-structured space to the historical anchor requires a much larger global rotation. While this requires more rotational effort, it ensures that the internal structure of the language remains intact, ultimately yielding vastly superior alignment precision.

\section{Evaluation of Semantic Drift}

\subsection{Statistical Neighbourhood Analysis} 
\label{app:SNA}

\begin{table}[h!tb]
\centering
\resizebox{0.98\columnwidth}{!}{
\begin{tabular}{lccccr}
\toprule
\textbf{Model} & \textbf{Approach} & \textbf{Comparisons} & \textbf{Flagged at $\mathbf{k=0.8}$} & \textbf{\% Flagged} & \textbf{Deviation from Theory} \\
\midrule

\multirow{3}{*}{\textbf{\texttt{Word2Vec}}} & SMA & 3,913 & 870 & 22.23\% & +1.04 pp \\
 & Analytical OP & 3,150 & 635 & 20.16\% & -1.03 pp \\
 & Neural OP & 3,150 & 648 & 20.57\% & -0.62 pp \\\hline
\multirow{3}{*}{\textbf{\texttt{FastText}}} & SMA & 3,755 & 628 & 16.72\% & -4.47 pp \\
& Analytical OP & 3,045 & 503 & 16.52\% & -4.67 pp \\
& Neural OP & 3,045 & 493 & 16.19\% & -5.00 pp \\

\midrule
Mean across combinations & & & & 18.73\% & -2.46 pp \\
Theoretical Gaussian ($\Phi(-0.8)$) & & & & 21.19\% & --- \\
\bottomrule
\small\textbf{Note:} pp = Percentage Point & & & & & \\
\end{tabular}}
\caption{Empirical validation of the $k=0.8$ drift-detection threshold against the theoretical Gaussian expectation across the SMA, Analytical OP and Neural OP approaches.}
\label{tab:k_threshold_validation}
\end{table}

An empirical examination of the choice of $k = 0.8$ as the drift-detection threshold multiplier (cutoff $= \mu - k\sigma$) used for this neighbourhood analysis is summarised in Table~\ref{tab:k_threshold_validation}. This validation addresses the question: does $k = 0.8$ serve as a stable, theoretically grounded proportion of the vocabulary, or is it merely an arbitrary, combination-specific artefact?

Under a Gaussian approximation of the similarity distribution, a cutoff of $\mu - 0.8\sigma$ is anticipated to flag the bottom $\Phi(-0.8) = 21.19\%$ of comparisons. The six combinations, \texttt{Word2Vec} and \texttt{FastText}, each evaluated under SMA, Analytical OP and Neural-OP alignment, flag between 16.19\% and 22.23\% of word/lemma instances at $k = 0.8$, yielding a mean of 18.73\% (with deviations ranging from $+1.04$ to $-5.00$ percentage points from the theoretical expectation). Every model-to-mapping combination remains within 5 percentage points of the theoretical value, despite the significant differences in the embedding model and alignment mechanism.

Two notable patterns emerge from the comparison of the six combinations. First, \texttt{FastText}'s three alignment approaches cluster remarkably tightly, with a range of only 0.53 percentage points (16.19\% to 16.72\%). This suggests that the proportion it flags at $k=0.8$ is primarily influenced by the similarity distribution within the embedding space, rather than the specific alignment mechanism employed. In contrast, \texttt{Word2Vec} demonstrates more variability across its three approaches, with a range of 2.07 percentage points (20.16\% to 22.23\%). Notably, the SMA approach, which assesses Pearson correlation between similarity profiles rather than the direct cosine similarity of aligned vectors, diverges somewhat from the two OP-based methods. Second, five out of the six observed proportions fall below the theoretical value of 21.19\%. \texttt{FastText} consistently displays a significant negative bias across all three of its approaches, with values ranging from $-4.47$ to $-5.00$ percentage points. This indicates that its empirical similarity distribution is more heavily concentrated on the low-similarity end than the Gaussian approximation would suggest. Conversely, \texttt{Word2Vec}'s proportions cluster closer to the theoretical value, with SMA even exceeding it marginally by $+1.04$ percentage points.

In summary, the consistency across combinations, with all six proportions falling within approximately 5 percentage points of the theoretical expectation despite spanning two embedding models and three distinct alignment mechanisms, reinforces the idea that $k=0.8$ is a principled and non-arbitrary choice. This value corresponds to a near-canonical and reproducible fraction of the vocabulary flagged as candidate drift, irrespective of the embedding model or alignment method used to generate the underlying similarity scores.

Each of these model-to-mapping combinations features 15 different century-specific alignments in word space. The frequency distribution of lemmas (used in the SMA approach) or words (used in the OP approaches), which show similarity and decline according to the details outlined in \S~\ref{sec:SNA}, is illustrated in Table~\ref{tab:dip_analysis}.

\begin{table*}[h!tb]
\centering
\resizebox{0.98\textwidth}{!}{
\begin{tabular}{l|l|r|r|r|r|r|r|r|r|r|r|r|r|r|r|r|r}
\hline
 & \textbf{Mapping} & \textbf{13$\leftrightarrow$14} & \textbf{13$\leftrightarrow$15} & \textbf{13$\leftrightarrow$18} & \textbf{13$\leftrightarrow$19} & \textbf{13$\leftrightarrow$20} & \textbf{14$\leftrightarrow$15} & \textbf{14$\leftrightarrow$18} & \textbf{14$\leftrightarrow$19} & \textbf{14$\leftrightarrow$20} & \textbf{15$\leftrightarrow$18} & \textbf{15$\leftrightarrow$19} & \textbf{15$\leftrightarrow$20} & \textbf{18$\leftrightarrow$19} & \textbf{18$\leftrightarrow$20} & \textbf{19$\leftrightarrow$20} & \textbf{Total} \\
\hline
\multirow[t]{3}{*}{\texttt{Word2Vec}} & SMA & 13 & 12 & 17 & 16 & 17 & 13 & 16 & 16 & 14 & 20 & 12 & 14 & 16 & 19 & 14 & 229 \\\cline{2-18}
 & Analytical OP & 39 & 39 & 45 & 43 & 43 & 37 & 35 & 38 & 43 & 37 & 48 & 46 & 49 & 46 & 47 & 635 \\\cline{2-18}
 & Neural OP & 37 & 47 & 43 & 47 & 38 & 39 & 43 & 43 & 49 & 39 & 47 & 46 & 47 & 39 & 44 & 648 \\ \hline
\multirow[t]{3}{*}{\texttt{FastText}} & SMA & 6 & 1 & 8 & 4 & 2 & 6 & 6 & 3 & 3 & 6 & 0 & 1 & 5 & 7 & 1 & 59 \\\cline{2-18}
 & Analytical OP & 17 & 19 & 18 & 12 & 13 & 19 & 17 & 17 & 22 & 20 & 18 & 15 & 23 & 19 & 22 & 271 \\\cline{2-18}
 & Neural OP & 16 & 19 & 18 & 12 & 14 & 18 & 19 & 19 & 22 & 21 & 19 & 15 & 20 & 21 & 21 & 274 \\
\hline
\end{tabular}
}
\caption[]{\label{tab:dip_analysis} Frequency distribution of lemmas\textbackslash words exhibiting significant similarity dips across century-wise aligned semantic spaces (\texttt{Word2Vec} vs. \texttt{FastText}).}
\end{table*}

Subsequently, we conducted a neighbourhood analysis of the lemmas or words that have undergone semantic drift, following the approach detailed in \S~\ref{sec:SNA}. It is important to note that this is not a Bag-of-Words (BoW) analysis, such as the one conducted by~\citet{jayatilleke2026sidiacv20sinhaladiachroniccorpus} on the same subset of \texttt{SiDiaC-v.2.0} considered in this study. This neighbourhood analysis focuses on the top 5 similar words that remain closest within the latent space. By examining the related meanings of neighbouring words, as described in~\citet{Soratha_Godage} and using an interactive Sinhala dictionary, Wahara\footnote{\scriptsize \url{http://crawler.nlpc.uom.lk/}}, we aimed to disambiguate the possible meanings of the lemmas\textbackslash words that have experienced semantic drift.

Based on the six model-to-mapping combinations mentioned, we examine three intriguing examples of semantic drift for each. For the \texttt{Word2Vec} embeddings using the SMA approach, the selected examples are `\raisebox{-0.5ex}{%
\includegraphics[height=1.4\fontcharht\font`\A]{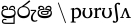}}', `\raisebox{-0.5ex}{%
\includegraphics[height=1.5\fontcharht\font`\A]{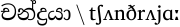}}', and `\raisebox{-0.5ex}{%
\includegraphics[height=1.3\fontcharht\font`\A]{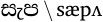}}', as shown in Table~\ref{tab:w2v_sma_ex}.

\begin{table*}[h!tb]
\centering
\resizebox{0.98\textwidth}{!}{
\begin{tabular}{c|c|c|c|c|c}
\hline
\textbf{Lemma} & \textbf{Similarity} & \textbf{Ref Century} & \textbf{Ref Century Neighbors} & \textbf{Target Century} & \textbf{Target Century Neighbors} \\
\hline
\raisebox{-0.5ex}{%
\includegraphics[height=1.4\fontcharht\font`\A]{Figures/si_008.pdf}} & 0.0274 & 18th & \makecell{\raisebox{-0.5ex}{%
\includegraphics[height=1.5\fontcharht\font`\A]{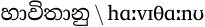}}, \raisebox{-0.5ex}{%
\includegraphics[height=1.4\fontcharht\font`\A]{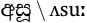}},\\ \raisebox{-0.5ex}{%
\includegraphics[height=1.5\fontcharht\font`\A]{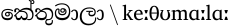}},\\ \raisebox{-0.5ex}{%
\includegraphics[height=1.3\fontcharht\font`\A]{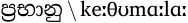}}, \raisebox{-0.5ex}{%
\includegraphics[height=1.5\fontcharht\font`\A]{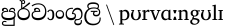}}} & 20th & \makecell{\raisebox{-0.5ex}{%
\includegraphics[height=1.3\fontcharht\font`\A]{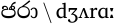}}, \raisebox{-0.5ex}{%
\includegraphics[height=1.3\fontcharht\font`\A]{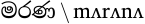}},\\ \raisebox{-0.5ex}{%
\includegraphics[height=1.5\fontcharht\font`\A]{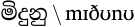}},\\ \raisebox{-0.5ex}{%
\includegraphics[height=1.5\fontcharht\font`\A]{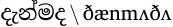}}, \raisebox{-0.5ex}{%
\includegraphics[height=1.5\fontcharht\font`\A]{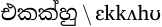}}} \\\hline
\raisebox{-0.5ex}{%
\includegraphics[height=1.5\fontcharht\font`\A]{Figures/si_009.pdf}} & 0.1379 & 14th & \makecell{\raisebox{-0.5ex}{%
\includegraphics[height=1.4\fontcharht\font`\A]{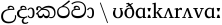}}, \raisebox{-0.5ex}{%
\includegraphics[height=1.5\fontcharht\font`\A]{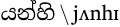}},\\ \raisebox{-0.5ex}{%
\includegraphics[height=1.5\fontcharht\font`\A]{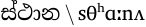}},\\ \raisebox{-0.5ex}{%
\includegraphics[height=1.4\fontcharht\font`\A]{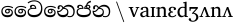}}, \raisebox{-0.5ex}{%
\includegraphics[height=1.4\fontcharht\font`\A]{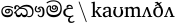}}} & 15th & \makecell{\raisebox{-0.5ex}{%
\includegraphics[height=1.5\fontcharht\font`\A]{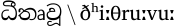}}, \raisebox{-0.5ex}{%
\includegraphics[height=1.5\fontcharht\font`\A]{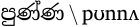}},\\ \raisebox{-0.5ex}{%
\includegraphics[height=1.3\fontcharht\font`\A]{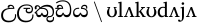}},\\ \raisebox{-0.5ex}{%
\includegraphics[height=1.4\fontcharht\font`\A]{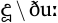}}, \raisebox{-0.5ex}{%
\includegraphics[height=1.5\fontcharht\font`\A]{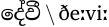}}} \\\hline
\raisebox{-0.5ex}{%
\includegraphics[height=1.3\fontcharht\font`\A]{Figures/si_010.pdf}} & 0.1755 & 14th & \makecell{\raisebox{-0.5ex}{%
\includegraphics[height=1.5\fontcharht\font`\A]{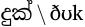}}, \raisebox{-0.5ex}{%
\includegraphics[height=1.4\fontcharht\font`\A]{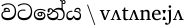}},\\ \raisebox{-0.5ex}{%
\includegraphics[height=1.4\fontcharht\font`\A]{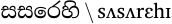}},\\ \raisebox{-0.5ex}{%
\includegraphics[height=1.3\fontcharht\font`\A]{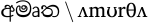}}, \raisebox{-0.5ex}{%
\includegraphics[height=1.5\fontcharht\font`\A]{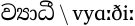}}} & 15th & \makecell{\raisebox{-0.5ex}{%
\includegraphics[height=1.3\fontcharht\font`\A]{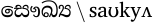}}, \raisebox{-0.5ex}{%
\includegraphics[height=1.4\fontcharht\font`\A]{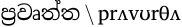}},\\ \raisebox{-0.5ex}{%
\includegraphics[height=1.4\fontcharht\font`\A]{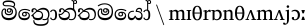}},\\ \raisebox{-0.5ex}{%
\includegraphics[height=1.3\fontcharht\font`\A]{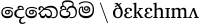}}, \raisebox{-0.5ex}{%
\includegraphics[height=1.4\fontcharht\font`\A]{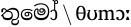}}} \\
\hline
\end{tabular}
}
\caption{\label{tab:w2v_sma_ex} Three lexical items exhibiting semantic drift, analysed using \texttt{Word2Vec} embeddings and Similarity Matrix Alignment (SMA).}
\end{table*}

The lemma `\raisebox{-0.5ex}{%
\includegraphics[height=1.4\fontcharht\font`\A]{Figures/si_008.pdf}}' in the 18th century was associated with descriptors highlighting the extraordinary physical attributes of the Supreme Buddha. Examples include `\raisebox{-0.5ex}{%
\includegraphics[height=1.4\fontcharht\font`\A]{Figures/si_013.pdf}}', which means the shining rays emanating from the Buddha's head, and `\raisebox{-0.5ex}{%
\includegraphics[height=1.3\fontcharht\font`\A]{Figures/si_012.pdf}}', referring to the number eighty, which relates to `\raisebox{-0.5ex}{%
\includegraphics[height=1.3\fontcharht\font`\A]{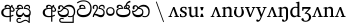}}\footnote{\scriptsize \url{https://pitaka.lk/books/pujawaliya/11-6.html}}', the eighty distinct physical characteristics of the Buddha. Additionally, the term `\raisebox{-0.5ex}{%
\includegraphics[height=1.3\fontcharht\font`\A]{Figures/si_014.pdf}}' signifies light, radiance, and brilliance, again connected to the physical attributes of the Buddha. By the 20th century, however, the word underwent a significant process of secularisation. It lost its ties to the supernatural attributes of the Buddha and instead began to reflect the ordinary human experience of mortality and suffering. Its semantic neighbours depict the raw realities of the ordinary individual (\raisebox{-0.5ex}{%
\includegraphics[height=1.4\fontcharht\font`\A]{Figures/si_020.pdf}}), bound to a deteriorating physical existence (\raisebox{-0.5ex}{%
\includegraphics[height=1.3\fontcharht\font`\A]{Figures/si_016.pdf}}) and death (\raisebox{-0.5ex}{%
\includegraphics[height=1.3\fontcharht\font`\A]{Figures/si_017.pdf}}).

The lemma `\raisebox{-0.5ex}{%
\includegraphics[height=1.5\fontcharht\font`\A]{Figures/si_009.pdf}}' usually means the moon. However, in the 14th century, it appears in a religious context where it is linked with words like `\raisebox{-0.5ex}{%
\includegraphics[height=1.4\fontcharht\font`\A]{Figures/si_021.pdf}}', which metaphorically refers to the spreading of the Buddha’s teachings as spiritual light across places. It also co-occurs with `\raisebox{-0.5ex}{%
\includegraphics[height=1.4\fontcharht\font`\A]{Figures/si_025.pdf}}', meaning a white lily or lotus, which is associated with purity and spirituality. By the 15th century, the meaning of `\raisebox{-0.5ex}{%
\includegraphics[height=1.5\fontcharht\font`\A]{Figures/si_009.pdf}}' changes. It moves away from religious use and becomes more connected to courtly and romantic language. In this period, it is often associated with royal and noble female terms such as `\raisebox{-0.5ex}{%
\includegraphics[height=1.4\fontcharht\font`\A]{Figures/si_026.pdf}}', `\raisebox{-0.5ex}{%
\includegraphics[height=1.4\fontcharht\font`\A]{Figures/si_029.pdf}}', and `\raisebox{-0.5ex}{%
\includegraphics[height=1.5\fontcharht\font`\A]{Figures/si_030.pdf}}', where the moon is used as a symbol of beautiful women and royal identity.

The core meaning of the lemma `\raisebox{-0.5ex}{%
\includegraphics[height=1.3\fontcharht\font`\A]{Figures/si_010.pdf}}' is happiness or comfort. In the 14th century, its closest semantic neighbours show a very spiritual and religious understanding. In this context, true happiness is not ordinary pleasure, but the complete ending of suffering in samsara, linked with words like \raisebox{-0.5ex}{%
\includegraphics[height=1.5\fontcharht\font`\A]{Figures/si_031.pdf}} and \raisebox{-0.5ex}{%
\includegraphics[height=1.3\fontcharht\font`\A]{Figures/si_033.pdf}}. It is also connected to `\raisebox{-0.5ex}{%
\includegraphics[height=1.2\fontcharht\font`\A]{Figures/si_034.pdf}}', meaning deathless Nibbāna, and `\raisebox{-0.5ex}{%
\includegraphics[height=1.4\fontcharht\font`\A]{Figures/si_035.pdf}}', meaning sickness or illness, which represents the suffering from which liberation is achieved. By the 15th century, the meaning of the lemma shifts away from this spiritual sense and becomes more connected to everyday life. Happiness is no longer understood in a spiritual manner, but is instead linked with physical well-being (\raisebox{-0.5ex}{%
\includegraphics[height=1.3\fontcharht\font`\A]{Figures/si_036.pdf}}) and warm social relationships (\raisebox{-0.5ex}{%
\includegraphics[height=1.4\fontcharht\font`\A]{Figures/si_038.pdf}}).

\begin{table*}[h!tb]
\centering
\resizebox{0.98\textwidth}{!}{
\begin{tabular}{c|c|c|c|c|c}
\hline
\textbf{Word} & \textbf{Similarity} & \textbf{Ref Century} & \textbf{Ref Century Neighbors} & \textbf{Target Century} & \textbf{Target Century Neighbors} \\
\hline
\raisebox{-0.5ex}{%
\includegraphics[height=1.4\fontcharht\font`\A]{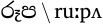}} & 0.4665 & 13th & \makecell{\raisebox{-0.5ex}{%
\includegraphics[height=1.5\fontcharht\font`\A]{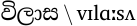}}, \raisebox{-0.5ex}{%
\includegraphics[height=1.4\fontcharht\font`\A]{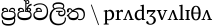}},\\ \raisebox{-0.5ex}{%
\includegraphics[height=1.3\fontcharht\font`\A]{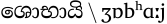}},\\ \raisebox{-0.5ex}{%
\includegraphics[height=1.5\fontcharht\font`\A]{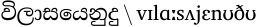}}, \raisebox{-0.5ex}{%
\includegraphics[height=1.4\fontcharht\font`\A]{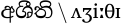}}} & 19th & \makecell{\raisebox{-0.5ex}{%
\includegraphics[height=1.4\fontcharht\font`\A]{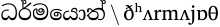}}, \raisebox{-0.5ex}{%
\includegraphics[height=1.3\fontcharht\font`\A]{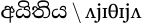}},\\ \raisebox{-0.5ex}{%
\includegraphics[height=1.5\fontcharht\font`\A]{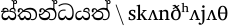}},\\ \raisebox{-0.5ex}{%
\includegraphics[height=1.4\fontcharht\font`\A]{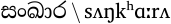}}, \raisebox{-0.5ex}{%
\includegraphics[height=1.4\fontcharht\font`\A]{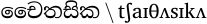}}} \\\hline
\raisebox{-0.5ex}{%
\includegraphics[height=1.4\fontcharht\font`\A]{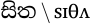}} & 0.4876 & 13th & \makecell{\raisebox{-0.5ex}{%
\includegraphics[height=1.5\fontcharht\font`\A]{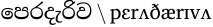}}, \raisebox{-0.5ex}{%
\includegraphics[height=1.5\fontcharht\font`\A]{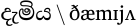}},\\ \raisebox{-0.5ex}{%
\includegraphics[height=1.4\fontcharht\font`\A]{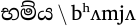}},\\ \raisebox{-0.5ex}{%
\includegraphics[height=1.5\fontcharht\font`\A]{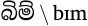}}, \raisebox{-0.5ex}{%
\includegraphics[height=1.3\fontcharht\font`\A]{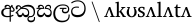}}} & 18th & \makecell{\raisebox{-0.5ex}{%
\includegraphics[height=1.5\fontcharht\font`\A]{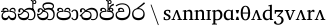}}, \raisebox{-0.5ex}{%
\includegraphics[height=1.5\fontcharht\font`\A]{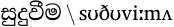}},\\ \raisebox{-0.5ex}{%
\includegraphics[height=1.35\fontcharht\font`\A]{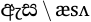}},\\ \raisebox{-0.5ex}{%
\includegraphics[height=1.5\fontcharht\font`\A]{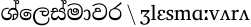}}, \raisebox{-0.5ex}{%
\includegraphics[height=1.5\fontcharht\font`\A]{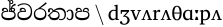}}} \\\hline
\raisebox{-0.5ex}{%
\includegraphics[height=1.5\fontcharht\font`\A]{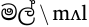}} & 0.5788 & 14th & \makecell{\raisebox{-0.5ex}{%
\includegraphics[height=1.5\fontcharht\font`\A]{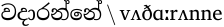}}, \raisebox{-0.5ex}{%
\includegraphics[height=1.3\fontcharht\font`\A]{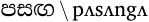}},\\ \raisebox{-0.5ex}{%
\includegraphics[height=1.5\fontcharht\font`\A]{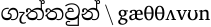}},\\ \raisebox{-0.5ex}{%
\includegraphics[height=1.4\fontcharht\font`\A]{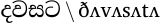}}, \raisebox{-0.5ex}{%
\includegraphics[height=1.3\fontcharht\font`\A]{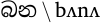}}} & 18th & \makecell{\raisebox{-0.5ex}{%
\includegraphics[height=1.5\fontcharht\font`\A]{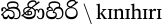}}, \raisebox{-0.5ex}{%
\includegraphics[height=1.5\fontcharht\font`\A]{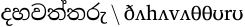}},\\ \raisebox{-0.5ex}{%
\includegraphics[height=1.4\fontcharht\font`\A]{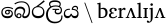}}, \\\raisebox{-0.5ex}{%
\includegraphics[height=1.4\fontcharht\font`\A]{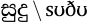}}, \raisebox{-0.5ex}{%
\includegraphics[height=1.4\fontcharht\font`\A]{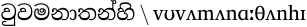}}} \\
\hline
\end{tabular}
}
\caption{\label{tab:w2v_opa_ex} Three lexical items exhibiting semantic drift, analysed using \texttt{Word2Vec} embeddings and Analytical Procrustes Alignment.}
\end{table*}

For the \texttt{Word2Vec} embeddings using the analytical procrustes alignment approach, the three selected examples are `\raisebox{-0.5ex}{%
\includegraphics[height=1.4\fontcharht\font`\A]{Figures/si_041.pdf}}', `\raisebox{-0.5ex}{%
\includegraphics[height=1.4\fontcharht\font`\A]{Figures/si_052.pdf}}', and `\raisebox{-0.5ex}{%
\includegraphics[height=1.4\fontcharht\font`\A]{Figures/si_001.pdf}}' as shown in Table~\ref{tab:w2v_opa_ex}.

In the 13th century, the word `\raisebox{-0.5ex}{%
\includegraphics[height=1.4\fontcharht\font`\A]{Figures/si_041.pdf}}' is associated with terms describing physical appearance and bodily form. Its semantic neighbours cluster around physical grace (\raisebox{-0.5ex}{%
\includegraphics[height=1.4\fontcharht\font`\A]{Figures/si_042.pdf}}, \raisebox{-0.5ex}{%
\includegraphics[height=1.4\fontcharht\font`\A]{Figures/si_043.pdf}}) and radiant visual beauty or splendour (\raisebox{-0.5ex}{%
\includegraphics[height=1.3\fontcharht\font`\A]{Figures/si_046.pdf}}), referring to brightness, brilliance, and aesthetic attractiveness. By the 19th century, the lemma shifts into a more technical and philosophical usage within Buddhist thought, where it denotes “form” or “matter.” In this later stage, its semantic neighbours belong to abstract doctrinal categories such as the five aggregates (\raisebox{-0.5ex}{%
\includegraphics[height=1.5\fontcharht\font`\A]{Figures/si_049.pdf}}, meaning the aggregates that constitute a being), \raisebox{-0.5ex}{%
\includegraphics[height=1.4\fontcharht\font`\A]{Figures/si_050.pdf}} (understood as conditioned phenomena encompassing all formations arising through karmic and causal processes, both wholesome and unwholesome), mental factors (\raisebox{-0.5ex}{%
\includegraphics[height=1.4\fontcharht\font`\A]{Figures/si_051.pdf}}), and the teachings of the Buddha (\raisebox{-0.5ex}{%
\includegraphics[height=1.5\fontcharht\font`\A]{Figures/si_047.pdf}}).

In the 13th century, the word `\raisebox{-0.5ex}{%
\includegraphics[height=1.4\fontcharht\font`\A]{Figures/si_052.pdf}}' is associated with a semantic field related to intention and basic mental activity. Its closest neighbours include terms such as prior intention or disposition (\raisebox{-0.5ex}{%
\includegraphics[height=1.5\fontcharht\font`\A]{Figures/si_053.pdf}}), action or engagement (\raisebox{-0.5ex}{%
\includegraphics[height=1.5\fontcharht\font`\A]{Figures/si_054.pdf}}), and ground or context (\raisebox{-0.5ex}{%
\includegraphics[height=1.5\fontcharht\font`\A]{Figures/si_056.pdf}}), along with unwholesome action (\raisebox{-0.5ex}{%
\includegraphics[height=1.3\fontcharht\font`\A]{Figures/si_057.pdf}}), suggesting an early connection between the mind and moral behaviour. By the 18th century, the lemma shifts into a more physical and pathological semantic field. Its neighbours include terms related to illness and bodily conditions, such as a type of fever involving accumulation of symptoms (\raisebox{-0.5ex}{%
\includegraphics[height=1.5\fontcharht\font`\A]{Figures/si_058.pdf}}), paleness (\raisebox{-0.5ex}{%
\includegraphics[height=1.5\fontcharht\font`\A]{Figures/si_060.pdf}}), eye (\raisebox{-0.5ex}{%
\includegraphics[height=1.35\fontcharht\font`\A]{Figures/si_061.pdf}}), phlegm-related condition (\raisebox{-0.5ex}{%
\includegraphics[height=1.5\fontcharht\font`\A]{Figures/si_062.pdf}}), and fever/heat (\raisebox{-0.5ex}{%
\includegraphics[height=1.5\fontcharht\font`\A]{Figures/si_063.pdf}}). In this period, `\raisebox{-0.5ex}{%
\includegraphics[height=1.4\fontcharht\font`\A]{Figures/si_052.pdf}}' is linked not only to mental processes but also to bodily states and health-related interpretations of experience.

In the 14th century, the word `\raisebox{-0.5ex}{%
\includegraphics[height=1.5\fontcharht\font`\A]{Figures/si_001.pdf}}' appears in a context related to religious listening and devotion. Its neighbours include speaking or preaching (\raisebox{-0.5ex}{%
\includegraphics[height=1.5\fontcharht\font`\A]{Figures/si_064.pdf}}), five ritual elements (\raisebox{-0.5ex}{%
\includegraphics[height=1.35\fontcharht\font`\A]{Figures/si_065.pdf}}), devotees (\raisebox{-0.5ex}{%
\includegraphics[height=1.5\fontcharht\font`\A]{Figures/si_066.pdf}}), and Buddhist preachings (\raisebox{-0.5ex}{%
\includegraphics[height=1.4\fontcharht\font`\A]{Figures/si_068.pdf}}). In this context, `\raisebox{-0.5ex}{%
\includegraphics[height=1.5\fontcharht\font`\A]{Figures/si_001.pdf}}' refers not only to flowers as physical objects, but also to their use as offerings in worship and in the setting of listening to Buddhist preachings. By the 18th century, the lemma shifts into a more natural and descriptive context. Its neighbours include plant names such as \raisebox{-0.5ex}{%
\includegraphics[height=1.5\fontcharht\font`\A]{Figures/si_069.pdf}} and \raisebox{-0.5ex}{%
\includegraphics[height=1.4\fontcharht\font`\A]{Figures/si_006.pdf}}, and the colour term `\raisebox{-0.5ex}{%
\includegraphics[height=1.4\fontcharht\font`\A]{Figures/si_005.pdf}}' (white). In this period, `\raisebox{-0.5ex}{%
\includegraphics[height=1.5\fontcharht\font`\A]{Figures/si_001.pdf}}' is mainly connected to flowers as part of the natural environment, rather than religious or ritual use.

\begin{table*}[h!tb]
\centering
\resizebox{0.98\textwidth}{!}{
\begin{tabular}{c|c|c|c|c|c}
\hline
\textbf{Word} & \textbf{Similarity} & \textbf{Ref Century} & \textbf{Ref Century Neighbors} & \textbf{Target Century} & \textbf{Target Century Neighbors} \\
\hline
\raisebox{-0.5ex}{%
\includegraphics[height=1.5\fontcharht\font`\A]{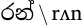}} & 0.2549 & 19th & \makecell{\raisebox{-0.5ex}{%
\includegraphics[height=1.5\fontcharht\font`\A]{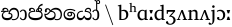}}, \raisebox{-0.5ex}{%
\includegraphics[height=1.5\fontcharht\font`\A]{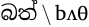}},\\ \raisebox{-0.5ex}{%
\includegraphics[height=1.3\fontcharht\font`\A]{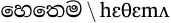}},\\ \raisebox{-0.5ex}{%
\includegraphics[height=1.3\fontcharht\font`\A]{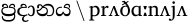}}, \raisebox{-0.5ex}{%
\includegraphics[height=1.4\fontcharht\font`\A]{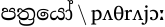}}} & 20th & \makecell{\raisebox{-0.5ex}{%
\includegraphics[height=1.5\fontcharht\font`\A]{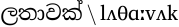}}, \raisebox{-0.5ex}{%
\includegraphics[height=1.5\fontcharht\font`\A]{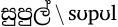}},\\ \raisebox{-0.5ex}{%
\includegraphics[height=1.4\fontcharht\font`\A]{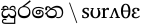}},\\ \raisebox{-0.5ex}{%
\includegraphics[height=1.5\fontcharht\font`\A]{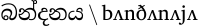}}, \raisebox{-0.5ex}{%
\includegraphics[height=1.5\fontcharht\font`\A]{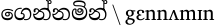}}} \\\hline
\raisebox{-0.5ex}{%
\includegraphics[height=1.5\fontcharht\font`\A]{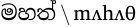}} & 0.2775 & 15th & \makecell{\raisebox{-0.5ex}{%
\includegraphics[height=1.5\fontcharht\font`\A]{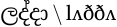}}, \raisebox{-0.5ex}{%
\includegraphics[height=1.5\fontcharht\font`\A]{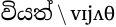}},\\ \raisebox{-0.5ex}{%
\includegraphics[height=1.5\fontcharht\font`\A]{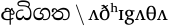}},\\ \raisebox{-0.5ex}{%
\includegraphics[height=1.5\fontcharht\font`\A]{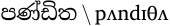}}, \raisebox{-0.5ex}{%
\includegraphics[height=1.3\fontcharht\font`\A]{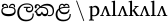}}} & 19th & \makecell{\raisebox{-0.5ex}{%
\includegraphics[height=1.3\fontcharht\font`\A]{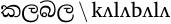}}, \raisebox{-0.5ex}{%
\includegraphics[height=1.4\fontcharht\font`\A]{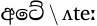}},\\ \raisebox{-0.5ex}{%
\includegraphics[height=1.3\fontcharht\font`\A]{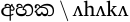}},\\ \raisebox{-0.5ex}{%
\includegraphics[height=1.5\fontcharht\font`\A]{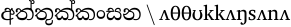}}, \raisebox{-0.5ex}{%
\includegraphics[height=1.5\fontcharht\font`\A]{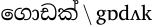}}} \\\hline
\raisebox{-0.5ex}{%
\includegraphics[height=1.5\fontcharht\font`\A]{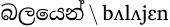}} & 0.371 & 15th & \makecell{\raisebox{-0.5ex}{%
\includegraphics[height=1.4\fontcharht\font`\A]{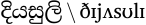}}, \raisebox{-0.5ex}{%
\includegraphics[height=1.4\fontcharht\font`\A]{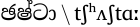}},\\ \raisebox{-0.5ex}{%
\includegraphics[height=1.4\fontcharht\font`\A]{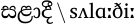}},\\ \raisebox{-0.5ex}{%
\includegraphics[height=1.4\fontcharht\font`\A]{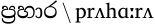}}, \raisebox{-0.5ex}{%
\includegraphics[height=1.5\fontcharht\font`\A]{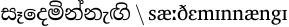}}} & 20th & \makecell{\raisebox{-0.5ex}{%
\includegraphics[height=1.5\fontcharht\font`\A]{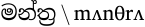}}, \raisebox{-0.5ex}{%
\includegraphics[height=1.5\fontcharht\font`\A]{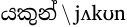}},\\ \raisebox{-0.5ex}{%
\includegraphics[height=1.4\fontcharht\font`\A]{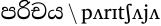}},\\ \raisebox{-0.5ex}{%
\includegraphics[height=1.4\fontcharht\font`\A]{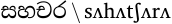}}, \raisebox{-0.5ex}{%
\includegraphics[height=1.4\fontcharht\font`\A]{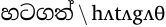}}} \\
\hline
\end{tabular}
}
\caption{\label{tab:w2v_npa_ex} Three lexical items exhibiting semantic drift, analysed using \texttt{Word2Vec} embeddings and Neural Procrustes Alignment.}
\end{table*}

For the \texttt{Word2Vec} embeddings using the neural procrustes alignment approach, the three selected examples are `\raisebox{-0.5ex}{%
\includegraphics[height=1.5\fontcharht\font`\A]{Figures/si_071.pdf}}', `\raisebox{-0.5ex}{%
\includegraphics[height=1.4\fontcharht\font`\A]{Figures/si_082.pdf}}', and `\raisebox{-0.5ex}{%
\includegraphics[height=1.5\fontcharht\font`\A]{Figures/si_093.pdf}}' as shown in Table~\ref{tab:w2v_npa_ex}.

In the 19th century, the word `\raisebox{-0.5ex}{%
\includegraphics[height=1.5\fontcharht\font`\A]{Figures/si_071.pdf}}' appears in a semantic environment connected with material objects and social actions. Its neighbours include containers or vessels (\raisebox{-0.5ex}{%
\includegraphics[height=1.5\fontcharht\font`\A]{Figures/si_072.pdf}}), food such as rice (\raisebox{-0.5ex}{%
\includegraphics[height=1.5\fontcharht\font`\A]{Figures/si_073.pdf}}), personal reference (\raisebox{-0.5ex}{%
\includegraphics[height=1.3\fontcharht\font`\A]{Figures/si_074.pdf}}), giving or offering (\raisebox{-0.5ex}{%
\includegraphics[height=1.3\fontcharht\font`\A]{Figures/si_075.pdf}}), and leaves or plates (\raisebox{-0.5ex}{%
\includegraphics[height=1.5\fontcharht\font`\A]{Figures/si_076.pdf}}). In this context, `\raisebox{-0.5ex}{%
\includegraphics[height=1.5\fontcharht\font`\A]{Figures/si_071.pdf}}' is associated with tangible items used in everyday or ritual transactions, especially objects involved in giving and offering. By the 20th century, the word shifts into a more poetic and ornamental semantic field. Its neighbours include decorative or natural elements such as a creeper or vine (\raisebox{-0.5ex}{%
\includegraphics[height=1.5\fontcharht\font`\A]{Figures/si_077.pdf}}), blooming or fullness (\raisebox{-0.5ex}{%
\includegraphics[height=1.5\fontcharht\font`\A]{Figures/si_078.pdf}}), the hand or palm (\raisebox{-0.5ex}{%
\includegraphics[height=1.4\fontcharht\font`\A]{Figures/si_079.pdf}}), binding or tying (\raisebox{-0.5ex}{%
\includegraphics[height=1.5\fontcharht\font`\A]{Figures/si_080.pdf}}), and bringing or leading (\raisebox{-0.5ex}{%
\includegraphics[height=1.5\fontcharht\font`\A]{Figures/si_081.pdf}}). In this period, `\raisebox{-0.5ex}{%
\includegraphics[height=1.5\fontcharht\font`\A]{Figures/si_071.pdf}}' is used more in an aesthetic and figurative sense, often linked with beauty, decoration, and poetic imagery rather than material exchange.

In the 15th century, the word `\raisebox{-0.5ex}{%
\includegraphics[height=1.5\fontcharht\font`\A]{Figures/si_082.pdf}}' appears in a semantic environment associated with intellectual and learned qualities. Its neighbours include terms such as attained or gained (\raisebox{-0.5ex}{%
\includegraphics[height=1.5\fontcharht\font`\A]{Figures/si_083.pdf}}), learned or educated (\raisebox{-0.5ex}{%
\includegraphics[height=1.5\fontcharht\font`\A]{Figures/si_084.pdf}}), realised or obtained (\raisebox{-0.5ex}{%
\includegraphics[height=1.5\fontcharht\font`\A]{Figures/si_085.pdf}}), scholar (\raisebox{-0.5ex}{%
\includegraphics[height=1.5\fontcharht\font`\A]{Figures/si_086.pdf}}), and published or expressed (\raisebox{-0.5ex}{%
\includegraphics[height=1.3\fontcharht\font`\A]{Figures/si_087.pdf}}). In this context, `\raisebox{-0.5ex}{%
\includegraphics[height=1.5\fontcharht\font`\A]{Figures/si_082.pdf}}' is linked with ideas of greatness in knowledge, learning, and intellectual achievement. By the 19th century, the word shifts into a more everyday and informal semantic field. Its neighbours include disturbance or confusion (\raisebox{-0.5ex}{%
\includegraphics[height=1.3\fontcharht\font`\A]{Figures/si_088.pdf}}), aside or away (\raisebox{-0.5ex}{%
\includegraphics[height=1.3\fontcharht\font`\A]{Figures/si_090.pdf}}), self-praise or arrogance (\raisebox{-0.5ex}{%
\includegraphics[height=1.5\fontcharht\font`\A]{Figures/si_091.pdf}}), and many or large quantities (\raisebox{-0.5ex}{%
\includegraphics[height=1.5\fontcharht\font`\A]{Figures/si_092.pdf}}). In this period, `\raisebox{-0.5ex}{%
\includegraphics[height=1.5\fontcharht\font`\A]{Figures/si_082.pdf}}' is used more in relation to quantity, intensity, or everyday descriptive meanings rather than scholarly or intellectual greatness.

In the 15th century, the word `\raisebox{-0.5ex}{%
\includegraphics[height=1.5\fontcharht\font`\A]{Figures/si_093.pdf}}' appears in a semantic environment associated with force and physical action. Its neighbours include whirlpool or swirling water (\raisebox{-0.5ex}{%
\includegraphics[height=1.4\fontcharht\font`\A]{Figures/si_094.pdf}}), swirling or rotational motion (\raisebox{-0.5ex}{%
\includegraphics[height=1.4\fontcharht\font`\A]{Figures/si_096.pdf}}), attack or strike (\raisebox{-0.5ex}{%
\includegraphics[height=1.4\fontcharht\font`\A]{Figures/si_097.pdf}}), and rising or emerging (\raisebox{-0.5ex}{%
\includegraphics[height=1.5\fontcharht\font`\A]{Figures/si_098.pdf}}). In this context, `\raisebox{-0.5ex}{%
\includegraphics[height=1.5\fontcharht\font`\A]{Figures/si_093.pdf}}' is linked with the idea of physical force, impact, and strong natural or dynamic motion. By the 20th century, the word shifted into a more supernatural and social semantic field. Its neighbours include spells or chants (\raisebox{-0.5ex}{%
\includegraphics[height=1.5\fontcharht\font`\A]{Figures/si_099.pdf}}), demons or spirits (\raisebox{-0.5ex}{%
\includegraphics[height=1.5\fontcharht\font`\A]{Figures/si_100.pdf}}), practice or experience (\raisebox{-0.5ex}{%
\includegraphics[height=1.4\fontcharht\font`\A]{Figures/si_101.pdf}}), companions or associates (\raisebox{-0.5ex}{%
\includegraphics[height=1.4\fontcharht\font`\A]{Figures/si_102.pdf}}), and arising or coming into being (\raisebox{-0.5ex}{%
\includegraphics[height=1.5\fontcharht\font`\A]{Figures/si_103.pdf}}). In this period, `\raisebox{-0.5ex}{%
\includegraphics[height=1.5\fontcharht\font`\A]{Figures/si_093.pdf}}' is associated more with supernatural power, ritual force, and socially embedded notions of influence rather than purely physical strength.

\begin{table*}[h!tb]
\centering
\resizebox{0.98\textwidth}{!}{
\begin{tabular}{c|c|c|c|c|c}
\hline
\textbf{Lemma} & \textbf{Similarity} & \textbf{Ref Century} & \textbf{Ref Century Neighbors} & \textbf{Target Century} & \textbf{Target Century Neighbors} \\
\hline
\raisebox{-0.5ex}{%
\includegraphics[height=1.3\fontcharht\font`\A]{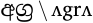}} & 0.2175 & 13th & \makecell{ \raisebox{-0.5ex}{%
\includegraphics[height=1.5\fontcharht\font`\A]{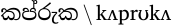}}, \raisebox{-0.5ex}{%
\includegraphics[height=1.5\fontcharht\font`\A]{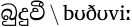}} , \\ \raisebox{-0.5ex}{%
\includegraphics[height=1.5\fontcharht\font`\A]{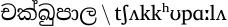}} ,  \\ \raisebox{-0.5ex}{%
\includegraphics[height=1.5\fontcharht\font`\A]{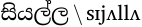}} ,\raisebox{-0.5ex}{%
\includegraphics[height=1.5\fontcharht\font`\A]{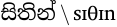}} } & 18th &  \makecell{\raisebox{-0.5ex}{%
\includegraphics[height=1.4\fontcharht\font`\A]{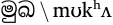}} , 
\raisebox{-0.5ex}{%
\includegraphics[height=1.4\fontcharht\font`\A]{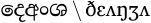}} , \\
\raisebox{-0.5ex}{%
\includegraphics[height=1.4\fontcharht\font`\A]{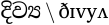}} , 
\raisebox{-0.5ex}{%
\includegraphics[height=1.5
\fontcharht\font`\A]{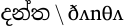}} , \\
\raisebox{-0.5ex}{%
\includegraphics[height=1.4\fontcharht\font`\A]{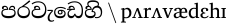}}
} \\\hline
\raisebox{-0.5ex}{%
\includegraphics[height=1.3\fontcharht\font`\A]{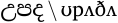}} & 0.3253 & 13th & \makecell{ \raisebox{-0.5ex}{%
\includegraphics[height=1.4\fontcharht\font`\A]{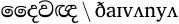}},  \raisebox{-0.5ex}{%
\includegraphics[height=1.4\fontcharht\font`\A]{Figures/si_017.pdf}},\\ \raisebox{-0.5ex}{%
\includegraphics[height=1.4\fontcharht\font`\A]{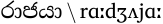}}, \\ \raisebox{-0.5ex}{%
\includegraphics[height=1.4\fontcharht\font`\A]{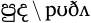}}, \raisebox{-0.5ex}{%
\includegraphics[height=1.5\fontcharht\font`\A]{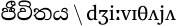}} } & 18th & 
\makecell{\raisebox{-0.5ex}{%
\includegraphics[height=1.5\fontcharht\font`\A]{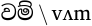}} ,
\raisebox{-0.5ex}{%
\includegraphics[height=1.35\fontcharht\font`\A]{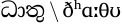}} , \\
\raisebox{-0.5ex}{%
\includegraphics[height=1.5\fontcharht\font`\A]{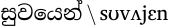}} , \\
\raisebox{-0.5ex}{%
\includegraphics[height=1.5\fontcharht\font`\A]{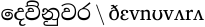}} ,
\raisebox{-0.5ex}{%
\includegraphics[height=1.5\fontcharht\font`\A]{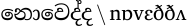}} }
\\\hline
\raisebox{-0.5ex}{%
\includegraphics[height=1.4\fontcharht\font`\A]{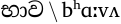}} & 0.4127 & 13th & \makecell{\raisebox{-0.5ex}{%
\includegraphics[height=1.5\fontcharht\font`\A]{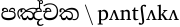}}, \raisebox{-0.5ex}{%
\includegraphics[height=1.5\fontcharht\font`\A]{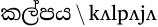}},\\ \raisebox{-0.5ex}{%
\includegraphics[height=1.4\fontcharht\font`\A]{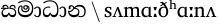}}, \\ \raisebox{-0.5ex}{%
\includegraphics[height=1.55\fontcharht\font`\A]{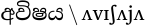}}, \raisebox{-0.5ex}{%
\includegraphics[height=1.55\fontcharht\font`\A]{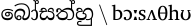}}} & 18th & \makecell{ \raisebox{-0.5ex}{%
\includegraphics[height=1.55\fontcharht\font`\A]{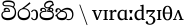}}, \raisebox{-0.5ex}{%
\includegraphics[height=1.55\fontcharht\font`\A]{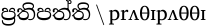}}, \\ \raisebox{-0.5ex}{%
\includegraphics[height=1.55\fontcharht\font`\A]{Figures/si_122.pdf}}, \\ \raisebox{-0.5ex}{%
\includegraphics[height=1.55\fontcharht\font`\A]{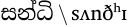}}, \raisebox{-0.5ex}{%
\includegraphics[height=1.55\fontcharht\font`\A]{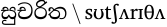}} }\\
\hline
\end{tabular}}
\caption{\label{tab:ft_sma_ex} Three lexical items exhibiting semantic drift, analysed using \texttt{FastText} embeddings and Similarity Matrix Alignment (SMA).}
\end{table*}

For the \texttt{FastText} embeddings using the SMA approach, the three selected examples are `\raisebox{-0.5ex}{%
\includegraphics[height=1.3\fontcharht\font`\A]{Figures/si_104.pdf}}', `\raisebox{-0.5ex}{%
\includegraphics[height=1.3\fontcharht\font`\A]{Figures/si_115.pdf}}', and `\raisebox{-0.5ex}{%
\includegraphics[height=1.4\fontcharht\font`\A]{Figures/si_123.pdf}}' as shown in Table~\ref{tab:ft_sma_ex}.

In the 13th century, the lemma `\raisebox{-0.5ex}{%
\includegraphics[height=1.3\fontcharht\font`\A]{Figures/si_104.pdf}}' appears in a semantic environment associated with spiritual and cosmological ideas. Its neighbors include the wish-fulfilling tree (\raisebox{-0.5ex}{%
\includegraphics[height=1.5\fontcharht\font`\A]{Figures/si_105.pdf}}), enlightenment or Buddhahood (\raisebox{-0.5ex}{%
\includegraphics[height=1.5\fontcharht\font`\A]{Figures/si_106.pdf}}), name of a monk (\raisebox{-0.5ex}{%
\includegraphics[height=1.5\fontcharht\font`\A]{Figures/si_107.pdf}}), universality or “all” (\raisebox{-0.5ex}{%
\includegraphics[height=1.5\fontcharht\font`\A]{Figures/si_108.pdf}}), and mental activity (\raisebox{-0.5ex}{%
\includegraphics[height=1.5\fontcharht\font`\A]{Figures/si_109.pdf}}). In this context, `\raisebox{-0.5ex}{%
\includegraphics[height=1.3\fontcharht\font`\A]{Figures/si_104.pdf}}' is linked with ideas of highest excellence, spiritual attainment, and supreme mental or religious states. By the 18th century, the lemma shifts into a more physical and bodily semantic field. Its neighbours include mouth or face (\raisebox{-0.5ex}{%
\includegraphics[height=1.4\fontcharht\font`\A]{Figures/si_110.pdf}}), two parts or divisions (\raisebox{-0.5ex}{%
\includegraphics[height=1.4\fontcharht\font`\A]{Figures/si_111.pdf}}), divine or celestial (\raisebox{-0.5ex}{%
\includegraphics[height=1.4\fontcharht\font`\A]{Figures/si_112.pdf}}), teeth (\raisebox{-0.5ex}{%
\includegraphics[height=1.5\fontcharht\font`\A]{Figures/si_113.pdf}}), and work or service for others (\raisebox{-0.5ex}{%
\includegraphics[height=1.4\fontcharht\font`\A]{Figures/si_114.pdf}}). In this period, `\raisebox{-0.5ex}{%
\includegraphics[height=1.3\fontcharht\font`\A]{Figures/si_104.pdf}}' is more associated with physical parts, bodily description, and social or divine attributes rather than purely spiritual or abstract excellence.

In the 13th century, the word `\raisebox{-0.5ex}{%
\includegraphics[height=1.3\fontcharht\font`\A]{Figures/si_115.pdf}}' appears in a semantic environment associated with life, destiny, and human existence. Its neighbors include astrologer or fortune teller (\raisebox{-0.5ex}{%
\includegraphics[height=1.4\fontcharht\font`\A]{Figures/si_167.pdf}}), death (\raisebox{-0.5ex}{%
\includegraphics[height=1.4\fontcharht\font`\A]{Figures/si_017.pdf}}), king (\raisebox{-0.5ex}{%
\includegraphics[height=1.4\fontcharht\font`\A]{Figures/si_116.pdf}}), worship or offering (\raisebox{-0.5ex}{%
\includegraphics[height=1.35\fontcharht\font`\A]{Figures/si_002.pdf}}), and life (\raisebox{-0.5ex}{%
\includegraphics[height=1.5\fontcharht\font`\A]{Figures/si_117.pdf}}). In this context, `\raisebox{-0.5ex}{%
\includegraphics[height=1.3\fontcharht\font`\A]{Figures/si_115.pdf}}' is connected with ideas of birth, fate, and the cycle of human existence. By the 18th century, the lemma shifts into a more physical and material semantic field. Its neighbors include left side (\raisebox{-0.5ex}{%
\includegraphics[height=1.5\fontcharht\font`\A]{Figures/si_118.pdf}}), bodily elements or substances (\raisebox{-0.5ex}{%
\includegraphics[height=1.35\fontcharht\font`\A]{Figures/si_119.pdf}}), well-being or comfort (\raisebox{-0.5ex}{%
\includegraphics[height=1.5\fontcharht\font`\A]{Figures/si_120.pdf}}), heavenly city or divine realm (\raisebox{-0.5ex}{%
\includegraphics[height=1.5\fontcharht\font`\A]{Figures/si_121.pdf}}), and negation or uncertainty (\raisebox{-0.5ex}{%
\includegraphics[height=1.5\fontcharht\font`\A]{Figures/si_122.pdf}}). In this period, “\raisebox{-0.5ex}{%
\includegraphics[height=1.3\fontcharht\font`\A]{Figures/si_115.pdf}}” becomes associated more with bodily, spatial, and worldly concepts rather than themes of destiny and existence.

In the 13th century, the lemma `\raisebox{-0.5ex}{%
\includegraphics[height=1.4\fontcharht\font`\A]{Figures/si_123.pdf}}' appears in a semantic environment associated with spiritual and philosophical ideas. Its neighbours include groups or sets of five (\raisebox{-0.5ex}{%
\includegraphics[height=1.5\fontcharht\font`\A]{Figures/si_168.pdf}}), cosmic age or aeon (\raisebox{-0.5ex}{%
\includegraphics[height=1.5\fontcharht\font`\A]{Figures/si_124.pdf}}), proper establishment or calming of the mind (\raisebox{-0.5ex}{%
\includegraphics[height=1.4\fontcharht\font`\A]{Figures/si_125.pdf}}), that which is beyond sensory perception or difficult to comprehend (\raisebox{-0.5ex}{%
\includegraphics[height=1.55\fontcharht\font`\A]{Figures/si_126.pdf}}), and Bodhisattvas (\raisebox{-0.5ex}{%
\includegraphics[height=1.55\fontcharht\font`\A]{Figures/si_127.pdf}}). In this context, `\raisebox{-0.5ex}{%
\includegraphics[height=1.4\fontcharht\font`\A]{Figures/si_123.pdf}}' is connected with religious states, spiritual development, and doctrinal thought. By the 18th century, the lemma shifts into a more ethical and practical semantic field. Its neighbours include radiant or brightly shining (\raisebox{-0.5ex}{%
\includegraphics[height=1.55\fontcharht\font`\A]{Figures/si_128.pdf}}), practice or conduct (\raisebox{-0.5ex}{%
\includegraphics[height=1.55\fontcharht\font`\A]{Figures/si_129.pdf}}), uncertainty or negation (\raisebox{-0.5ex}{%
\includegraphics[height=1.55\fontcharht\font`\A]{Figures/si_122.pdf}}), bodily joints or points of connection (\raisebox{-0.5ex}{%
\includegraphics[height=1.55\fontcharht\font`\A]{Figures/si_130.pdf}}), and good character or proper behaviour (\raisebox{-0.5ex}{%
\includegraphics[height=1.55\fontcharht\font`\A]{Figures/si_131.pdf}},). In this period, `\raisebox{-0.5ex}{%
\includegraphics[height=1.4\fontcharht\font`\A]{Figures/si_123.pdf}}' becomes more associated with moral behaviour, conduct, and practical human qualities rather than abstract spiritual or cosmological ideas.

\begin{table*}[h!tb]
\centering
\resizebox{0.98\textwidth}{!}{
\begin{tabular}{c|c|c|c|c|c}
\hline
\textbf{Word} & \textbf{Similarity} & \textbf{Ref Century} & \textbf{Ref Century Neighbors} & \textbf{Target Century} & \textbf{Target Century Neighbors} \\
\hline
\raisebox{-0.5ex}{%
\includegraphics[height=1.5\fontcharht\font`\A]{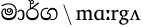}} & 0.6561 & 13th & \makecell{ \raisebox{-0.5ex}{%
\includegraphics[height=1.5\fontcharht\font`\A]{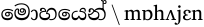}}, \raisebox{-0.5ex}{%
\includegraphics[height=1.4\fontcharht\font`\A]{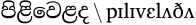}}, \\ \raisebox{-0.5ex}{%
\includegraphics[height=1.55\fontcharht\font`\A]{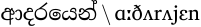}}, \\ \raisebox{-0.5ex}{%
\includegraphics[height=1.5\fontcharht\font`\A]{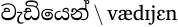}}, \raisebox{-0.5ex}{%
\includegraphics[height=1.4\fontcharht\font`\A]{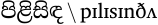}} } & 18th & \makecell{ \raisebox{-0.5ex}{%
\includegraphics[height=1.5\fontcharht\font`\A]{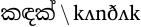}}, \raisebox{-0.5ex}{%
\includegraphics[height=1.5\fontcharht\font`\A]{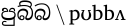}}, \\ \raisebox{-0.5ex}{%
\includegraphics[height=1.4\fontcharht\font`\A]{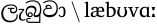}}, \\ \raisebox{-0.5ex}{%
\includegraphics[height=1.4\fontcharht\font`\A]{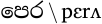}}, \raisebox{-0.5ex}{%
\includegraphics[height=1.5\fontcharht\font`\A]{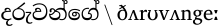}}}  \\\hline
\raisebox{-0.5ex}{%
\includegraphics[height=1.4\fontcharht\font`\A]{Figures/si_141.pdf}} & 0.7222 & 14th & \makecell{ \raisebox{-0.5ex}{%
\includegraphics[height=1.3\fontcharht\font`\A]{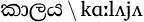}}, \raisebox{-0.5ex}{%
\includegraphics[height=1.5\fontcharht\font`\A]{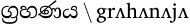}}, \\ \raisebox{-0.5ex}{%
\includegraphics[height=1.5\fontcharht\font`\A]{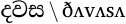}},\\  \raisebox{-0.5ex}{%
\includegraphics[height=1.5\fontcharht\font`\A]{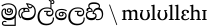}}, \raisebox{-0.5ex}{%
\includegraphics[height=1.5\fontcharht\font`\A]{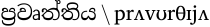}} } & 15th & \makecell{ \raisebox{-0.5ex}{%
\includegraphics[height=1.3\fontcharht\font`\A]{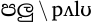}}, \raisebox{-0.5ex}{%
\includegraphics[height=1.3\fontcharht\font`\A]{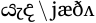}}, \\ \raisebox{-0.5ex}{%
\includegraphics[height=1.55\fontcharht\font`\A]{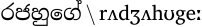}}, \\ \raisebox{-0.5ex}{%
\includegraphics[height=1.4\fontcharht\font`\A]{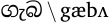}}, \raisebox{-0.5ex}{%
\includegraphics[height=1.5\fontcharht\font`\A]{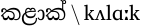}}} \\\hline
\raisebox{-0.5ex}{%
\includegraphics[height=1.5\fontcharht\font`\A]{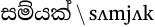}} & 0.7779 & 14th & \makecell{ \raisebox{-0.5ex}{%
\includegraphics[height=1.5\fontcharht\font`\A]{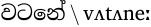}}, \raisebox{-0.5ex}{%
\includegraphics[height=1.4\fontcharht\font`\A]{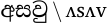}}, \\ \raisebox{-0.5ex}{%
\includegraphics[height=1.5\fontcharht\font`\A]{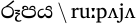}},\\   \raisebox{-0.5ex}{%
\includegraphics[height=1.5\fontcharht\font`\A]{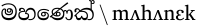}}, \raisebox{-0.5ex}{%
\includegraphics[height=1.5\fontcharht\font`\A]{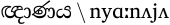}}} & 20th & \makecell{ \raisebox{-0.5ex}{%
\includegraphics[height=1.5\fontcharht\font`\A]{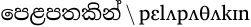}}, \raisebox{-0.5ex}{%
\includegraphics[height=1.5\fontcharht\font`\A]{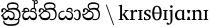}}, \\ \raisebox{-0.5ex}{%
\includegraphics[height=1.5\fontcharht\font`\A]{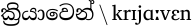}}, \\ \raisebox{-0.5ex}{%
\includegraphics[height=1.5\fontcharht\font`\A]{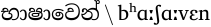}}, \raisebox{-0.5ex}{%
\includegraphics[height=1.5\fontcharht\font`\A]{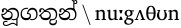}} } \\
\hline
\end{tabular}}
\caption{\label{tab:ft_opa_ex} Three lexical items exhibiting semantic drift, analysed using \texttt{FastText} embeddings and Orthogonal Procrustes Alignment.}
\end{table*}

For the \texttt{FastText} embeddings using the analytical procrustes alignment approach, the three selected examples are `\raisebox{-0.5ex}{%
\includegraphics[height=1.5\fontcharht\font`\A]{Figures/si_132.pdf}}', `\raisebox{-0.5ex}{%
\includegraphics[height=1.4\fontcharht\font`\A]{Figures/si_141.pdf}}', and `\raisebox{-0.5ex}{%
\includegraphics[height=1.5\fontcharht\font`\A]{Figures/si_151.pdf}}' as shown in Table~\ref{tab:ft_opa_ex}.

In the 13th century, the word `\raisebox{-0.5ex}{%
\includegraphics[height=1.5\fontcharht\font`\A]{Figures/si_132.pdf}}' appears in a semantic environment associated with guidance, order, and progression. Its neighbours include in delusion or ignorance (\raisebox{-0.5ex}{%
\includegraphics[height=1.5\fontcharht\font`\A]{Figures/si_133.pdf}}), order or arrangement (\raisebox{-0.5ex}{%
\includegraphics[height=1.4\fontcharht\font`\A]{Figures/si_134.pdf}}), with affection or love (\raisebox{-0.5ex}{%
\includegraphics[height=1.55\fontcharht\font`\A]{Figures/si_135.pdf}}), in greater degree (\raisebox{-0.5ex}{%
\includegraphics[height=1.5\fontcharht\font`\A]{Figures/si_136.pdf}}), and conception or arising (\raisebox{-0.5ex}{%
\includegraphics[height=1.4\fontcharht\font`\A]{Figures/si_137.pdf}}). In this context, `\raisebox{-0.5ex}{%
\includegraphics[height=1.5\fontcharht\font`\A]{Figures/si_132.pdf}}' is linked with ideas of a path as a structured way of conduct, moral direction, or spiritual progression. By the 18th century, the word shifts into a more physical and textual semantic field. Its neighbours include the main body of a person excluding the head and limbs (\raisebox{-0.5ex}{%
\includegraphics[height=1.5\fontcharht\font`\A]{Figures/si_138.pdf}}), a section or division of a book or collection of texts (\raisebox{-0.5ex}{%
\includegraphics[height=1.5\fontcharht\font`\A]{Figures/si_139.pdf}}), received or obtained (\raisebox{-0.5ex}{%
\includegraphics[height=1.4\fontcharht\font`\A]{Figures/si_140.pdf}}), before (\raisebox{-0.5ex}{%
\includegraphics[height=1.4\fontcharht\font`\A]{Figures/si_141.pdf}}), and children (\raisebox{-0.5ex}{%
\includegraphics[height=1.5\fontcharht\font`\A]{Figures/si_142.pdf}}). In this period, `\raisebox{-0.5ex}{%
\includegraphics[height=1.5\fontcharht\font`\A]{Figures/si_132.pdf}}' is more associated with physical movement, textual organisation, and everyday relational contexts rather than abstract moral or spiritual pathways.

In the 14th century, the word `\raisebox{-0.5ex}{%
\includegraphics[height=1.4\fontcharht\font`\A]{Figures/si_141.pdf}}' appears in a semantic environment associated with time and temporal reference. Its neighbors include time (\raisebox{-0.5ex}{%
\includegraphics[height=1.3\fontcharht\font`\A]{Figures/si_143.pdf}}), grasping or understanding (\raisebox{-0.5ex}{%
\includegraphics[height=1.4\fontcharht\font`\A]{Figures/si_144.pdf}}), day (\raisebox{-0.5ex}{%
\includegraphics[height=1.4\fontcharht\font`\A]{Figures/si_145.pdf}}), throughout or entirely (\raisebox{-0.5ex}{%
\includegraphics[height=1.5\fontcharht\font`\A]{Figures/si_177.pdf}}), and news or event (\raisebox{-0.5ex}{%
\includegraphics[height=1.5\fontcharht\font`\A]{Figures/si_178.pdf}}). In this context, `\raisebox{-0.5ex}{%
\includegraphics[height=1.4\fontcharht\font`\A]{Figures/si_141.pdf}}' is linked with ideas of time, sequence, and temporal framing of events. By the 15th century, the word shifts into a more narrative and situational semantic field. Its neighbours include fruit or leaf (\raisebox{-0.5ex}{%
\includegraphics[height=1.3\fontcharht\font`\A]{Figures/si_146.pdf}}), invited or requested through prayer (\raisebox{-0.5ex}{%
\includegraphics[height=1.3\fontcharht\font`\A]{Figures/si_147.pdf}}), of the king (\raisebox{-0.5ex}{%
\includegraphics[height=1.55\fontcharht\font`\A]{Figures/si_148.pdf}}), womb or inner space (\raisebox{-0.5ex}{%
\includegraphics[height=1.4\fontcharht\font`\A]{Figures/si_149.pdf}}), and done or made (\raisebox{-0.5ex}{%
\includegraphics[height=1.5\fontcharht\font`\A]{Figures/si_150.pdf}}). In this period, `\raisebox{-0.5ex}{%
\includegraphics[height=1.4\fontcharht\font`\A]{Figures/si_141.pdf}}' is used more within storytelling contexts and event descriptions rather than purely as a temporal marker.

In the 14th century, the word `\raisebox{-0.5ex}{%
\includegraphics[height=1.5\fontcharht\font`\A]{Figures/si_151.pdf}}' appears in a semantic environment associated with correctness, knowledge, and spiritual discipline. Its neighbours include worthy or valuable (\raisebox{-0.5ex}{%
\includegraphics[height=1.5\fontcharht\font`\A]{Figures/si_152.pdf}}), hearing or listening (\raisebox{-0.5ex}{%
\includegraphics[height=1.4\fontcharht\font`\A]{Figures/si_153.pdf}}), form or appearance (\raisebox{-0.5ex}{%
\includegraphics[height=1.4\fontcharht\font`\A]{Figures/si_154.pdf}}), knowledge or insight (\raisebox{-0.5ex}{%
\includegraphics[height=1.5\fontcharht\font`\A]{Figures/si_179.pdf}}), and a monk (\raisebox{-0.5ex}{%
\includegraphics[height=1.5\fontcharht\font`\A]{Figures/si_155.pdf}}). In this context, `\raisebox{-0.5ex}{%
\includegraphics[height=1.5\fontcharht\font`\A]{Figures/si_151.pdf}}' is linked with ideas of right understanding, proper conduct, and spiritual or religious correctness. By the 20th century, the word shifted into a more social and cultural semantic field. Its neighbors include lineage or descent (\raisebox{-0.5ex}{%
\includegraphics[height=1.5\fontcharht\font`\A]{Figures/si_156.pdf}}), Christian (\raisebox{-0.5ex}{%
\includegraphics[height=1.5\fontcharht\font`\A]{Figures/si_157.pdf}}), through action (\raisebox{-0.5ex}{%
\includegraphics[height=1.5\fontcharht\font`\A]{Figures/si_158.pdf}}), through language (\raisebox{-0.5ex}{%
\includegraphics[height=1.5\fontcharht\font`\A]{Figures/si_159.pdf}}), and uneducated people (\raisebox{-0.5ex}{%
\includegraphics[height=1.5\fontcharht\font`\A]{Figures/si_160.pdf}}). In this period, `\raisebox{-0.5ex}{%
\includegraphics[height=1.5\fontcharht\font`\A]{Figures/si_151.pdf}}' is used more in relation to social identity, cultural practices, and modes of expression rather than strictly religious or doctrinal correctness.

\begin{table*}[h!tb]
\centering
\resizebox{0.98\textwidth}{!}{
\begin{tabular}{c|c|c|c|c|c}
\hline
\textbf{Word} & \textbf{Similarity} & \textbf{Ref Century} & \textbf{Ref Century Neighbors} & \textbf{Target Century} & \textbf{Target Century Neighbors} \\
\hline
\raisebox{-0.5ex}{%
\includegraphics[height=1.5\fontcharht\font`\A]{Figures/si_161.pdf}} & 0.7112 & 14th & \makecell{
\raisebox{-0.5ex}{%
\includegraphics[height=1.5\fontcharht\font`\A]{Figures/si_162.pdf}}, \raisebox{-0.5ex}{%
\includegraphics[height=1.4\fontcharht\font`\A]{Figures/si_163.pdf}}, \\
\raisebox{-0.5ex}{%
\includegraphics[height=1.5\fontcharht\font`\A]{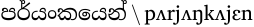}}, \\
\raisebox{-0.5ex}{%
\includegraphics[height=1.4\fontcharht\font`\A]{Figures/si_165.pdf}},
\raisebox{-0.5ex}{%
\includegraphics[height=1.5\fontcharht\font`\A]{Figures/si_166.pdf}} } & 15th & \makecell{ \raisebox{-0.5ex}{%
\includegraphics[height=1.5\fontcharht\font`\A]{Figures/si_169.pdf}}, \raisebox{-0.5ex}{%
\includegraphics[height=1.4\fontcharht\font`\A]{Figures/si_170.pdf}}, \\ \raisebox{-0.5ex}{%
\includegraphics[height=1.5\fontcharht\font`\A]{Figures/si_171.pdf}}, \\ \raisebox{-0.5ex}{%
\includegraphics[height=1.5\fontcharht\font`\A]{Figures/si_172.pdf}}, \raisebox{-0.5ex}{%
\includegraphics[height=1.5\fontcharht\font`\A]{Figures/si_173.pdf}} } \\\hline
\raisebox{-0.5ex}{%
\includegraphics[height=1.4\fontcharht\font`\A]{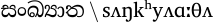}} & 0.7178 & 13th & \makecell{ \raisebox{-0.5ex}{%
\includegraphics[height=1.3\fontcharht\font`\A]{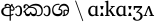}}, \raisebox{-0.5ex}{%
\includegraphics[height=1.5\fontcharht\font`\A]{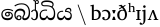}}, \\ \raisebox{-0.5ex}{%
\includegraphics[height=1.4\fontcharht\font`\A]{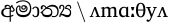}}, \\ \raisebox{-0.5ex}{%
\includegraphics[height=1.5\fontcharht\font`\A]{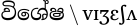}}, \raisebox{-0.5ex}{%
\includegraphics[height=1.4\fontcharht\font`\A]{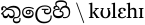}}}  & 19th & \makecell{ \raisebox{-0.5ex}{%
\includegraphics[height=1.5\fontcharht\font`\A]{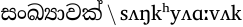}}, \raisebox{-0.5ex}{%
\includegraphics[height=1.5\fontcharht\font`\A]{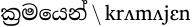}}, \\ \raisebox{-0.5ex}{%
\includegraphics[height=1.5\fontcharht\font`\A]{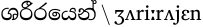}}, \\ \raisebox{-0.5ex}{%
\includegraphics[height=1.5\fontcharht\font`\A]{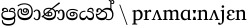}}, \raisebox{-0.5ex}{%
\includegraphics[height=1.4\fontcharht\font`\A]{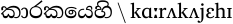}} } \\\hline
\raisebox{-0.5ex}{%
\includegraphics[height=1.4\fontcharht\font`\A]{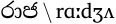}} & 0.7335 & 19th & \makecell{\raisebox{-0.5ex}{%
\includegraphics[height=1.3\fontcharht\font`\A]{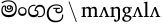}},
\raisebox{-0.5ex}{%
\includegraphics[height=1.3\fontcharht\font`\A]{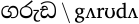}},\\
\raisebox{-0.5ex}{%
\includegraphics[height=1.4\fontcharht\font`\A]{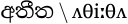}}, \\
\raisebox{-0.5ex}{%
\includegraphics[height=1.55\fontcharht\font`\A]{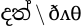}}, 
\raisebox{-0.5ex}{%
\includegraphics[height=1.5\fontcharht\font`\A]{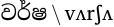}}} & 20th & \makecell {\raisebox{-0.5ex}{%
\includegraphics[height=1.4\fontcharht\font`\A]{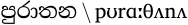}},  \raisebox{-0.5ex}{%
\includegraphics[height=1.5\fontcharht\font`\A]{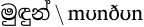}}, \\ \raisebox{-0.5ex}{%
\includegraphics[height=1.5\fontcharht\font`\A]{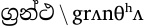}}, \\ \raisebox{-0.5ex}{%
\includegraphics[height=1.3\fontcharht\font`\A]{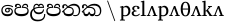}}, \raisebox{-0.5ex}{%
\includegraphics[height=1.5\fontcharht\font`\A]{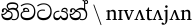}} } \\
\hline
\end{tabular}}
\caption{\label{tab:ft_npa_ex} Three lexical items exhibiting semantic drift, analysed using \texttt{FastText} embeddings and Neural Procrustes Alignment.}
\end{table*}

For the \texttt{FastText} embeddings using the neural procrustes alignment approach, the three selected examples are `\raisebox{-0.5ex}{%
\includegraphics[height=1.5\fontcharht\font`\A]{Figures/si_161.pdf}}', `\raisebox{-0.5ex}{%
\includegraphics[height=1.4\fontcharht\font`\A]{Figures/si_174.pdf}}', and `\raisebox{-0.5ex}{%
\includegraphics[height=1.4\fontcharht\font`\A]{Figures/si_188.pdf}}' as shown in Table~\ref{tab:ft_npa_ex}. The semantic change of the word `\raisebox{-0.5ex}{%
\includegraphics[height=1.5\fontcharht\font`\A]{Figures/si_161.pdf}}' over the two centuries was discussed in \S~\ref{sec:SNA}.

In the 13th century, the word `\raisebox{-0.5ex}{%
\includegraphics[height=1.4\fontcharht\font`\A]{Figures/si_174.pdf}}' appears in a semantic environment associated with abstract, spiritual, and social concepts. Its neighbours include sky or space (\raisebox{-0.5ex}{%
\includegraphics[height=1.3\fontcharht\font`\A]{Figures/si_175.pdf}}), enlightenment or awakening (\raisebox{-0.5ex}{%
\includegraphics[height=1.5\fontcharht\font`\A]{Figures/si_176.pdf}}), minister or official (\raisebox{-0.5ex}{%
\includegraphics[height=1.4\fontcharht\font`\A]{Figures/si_180.pdf}}), special or distinguished (\raisebox{-0.5ex}{%
\includegraphics[height=1.5\fontcharht\font`\A]{Figures/si_181.pdf}}), and clan or lineage (\raisebox{-0.5ex}{%
\includegraphics[height=1.4\fontcharht\font`\A]{Figures/si_182.pdf}}). In this context, `\raisebox{-0.5ex}{%
\includegraphics[height=1.4\fontcharht\font`\A]{Figures/si_174.pdf}}' is linked with ideas of higher status, spiritual significance, and structured social or cosmological order. By the 19th century, the word shifts into a more quantitative and descriptive semantic field. Its neighbors include number (\raisebox{-0.5ex}{%
\includegraphics[height=1.5\fontcharht\font`\A]{Figures/si_183.pdf}}), gradually or in sequence (\raisebox{-0.5ex}{%
\includegraphics[height=1.5\fontcharht\font`\A]{Figures/si_184.pdf}}), from the body (\raisebox{-0.5ex}{%
\includegraphics[height=1.5\fontcharht\font`\A]{Figures/si_185.pdf}}), by amount or measure (\raisebox{-0.5ex}{%
\includegraphics[height=1.5\fontcharht\font`\A]{Figures/si_186.pdf}}), and in relation to an agent or doer (\raisebox{-0.5ex}{%
\includegraphics[height=1.3\fontcharht\font`\A]{Figures/si_187.pdf}}). In this period, `\raisebox{-0.5ex}{%
\includegraphics[height=1.4\fontcharht\font`\A]{Figures/si_174.pdf}}' is more associated with numerical expression, measurement, and structured description rather than abstract or spiritual associations.

In the 19th century, the word `\raisebox{-0.5ex}{%
\includegraphics[height=1.4\fontcharht\font`\A]{Figures/si_188.pdf}}' appears in a semantic environment associated with royal and ceremonial concepts. Its neighbours include auspicious or ceremonial (\raisebox{-0.5ex}{%
\includegraphics[height=1.3\fontcharht\font`\A]{Figures/si_189.pdf}}), a mythical bird considered the supreme enemy of snakes (\raisebox{-0.5ex}{%
\includegraphics[height=1.3\fontcharht\font`\A]{Figures/si_190.pdf}}), past or ancient (\raisebox{-0.5ex}{%
\includegraphics[height=1.4\fontcharht\font`\A]{Figures/si_191.pdf}}), teeth or wisdom-related imagery (\raisebox{-0.5ex}{%
\includegraphics[height=1.55\fontcharht\font`\A]{Figures/si_192.pdf}}), and year or time (\raisebox{-0.5ex}{%
\includegraphics[height=1.5\fontcharht\font`\A]{Figures/si_193.pdf}}). In this context, “\raisebox{-0.5ex}{%
\includegraphics[height=1.4\fontcharht\font`\A]{Figures/si_188.pdf}}” is linked with ideas of kingship, tradition, and symbolic or ritual associations. By the 20th century, the word shifted into a more historical and textual semantic field. Its neighbours include ancient or traditional (\raisebox{-0.5ex}{%
\includegraphics[height=1.4\fontcharht\font`\A]{Figures/si_194.pdf}}), crown or top/head (\raisebox{-0.5ex}{%
\includegraphics[height=1.5\fontcharht\font`\A]{Figures/si_195.pdf}}), book or manuscript (\raisebox{-0.5ex}{%
\includegraphics[height=1.5\fontcharht\font`\A]{Figures/si_196.pdf}}), lineage or genealogy (\raisebox{-0.5ex}{%
\includegraphics[height=1.3\fontcharht\font`\A]{Figures/si_197.pdf}}), and weak or low-status persons (\raisebox{-0.5ex}{%
\includegraphics[height=1.5\fontcharht\font`\A]{Figures/si_198.pdf}}). In this period, `\raisebox{-0.5ex}{%
\includegraphics[height=1.4\fontcharht\font`\A]{Figures/si_188.pdf}}' is more associated with historical identity, textual tradition, and social hierarchy rather than active royal or ceremonial power.

\subsection{Bidirectional Semantic Impact Scoring}
\label{app:BSIS}

\begin{table*}[h!tb]
\centering
\resizebox{0.9\textwidth}{!}{
\begin{tabular}{c|c|c|c|c|c}
\hline
\textbf{Permutation} & \textbf{Avg Global Distance} & \textbf{Avg Persistent Shift} & \textbf{Avg Drift Reduction} \% & \textbf{Drifted Instances} & \textbf{Total Instances} \\
\hline
$13\rightarrow14$ & 0.0219 & 0.0176 & 12.6893 & 394 & 1948 \\
$13\rightarrow15$ & 0.0165 & 0.0139 & 14.8558 & 509 & 2527 \\
$13\rightarrow18$ & 0.0259 & 0.0215 & 13.5800 & 381 & 1591 \\
$13\rightarrow19$ & 0.0169 & 0.0142 & 13.9794 & 849 & 4653 \\
$13\rightarrow20$ & 0.0164 & 0.0140 & 17.7201 & 1077 & 5842 \\
$14\rightarrow13$ & 0.0256 & 0.0220 & 16.5676 & 1004 & 4875 \\
$14\rightarrow15$ & 0.0277 & 0.0238 & 15.8754 & 580 & 2580 \\
$14\rightarrow18$ & 0.0331 & 0.0279 & 12.7873 & 439 & 1663 \\
$14\rightarrow19$ & 0.0260 & 0.0220 & 17.4911 & 973 & 4658 \\
$14\rightarrow20$ & 0.0281 & 0.0246 & 14.7832 & 1231 & 5880 \\
$15\rightarrow13$ & 0.0184 & 0.0158 & 15.8875 & 917 & 4835 \\
$15\rightarrow14$ & 0.0264 & 0.0220 & 6.9842 & 427 & 2001 \\
$15\rightarrow18$ & 0.0264 & 0.0225 & 14.5676 & 374 & 1642 \\
$15\rightarrow19$ & 0.0228 & 0.0191 & 16.3489 & 886 & 4709 \\
$15\rightarrow20$ & 0.0194 & 0.0168 & 14.8761 & 1092 & 5873 \\
$18\rightarrow13$ & 0.0311 & 0.0277 & 13.3125 & 906 & 4864 \\
$18\rightarrow14$ & 0.0334 & 0.0292 & 13.4692 & 436 & 2015 \\
$18\rightarrow15$ & 0.0326 & 0.0288 & 13.5372 & 570 & 2630 \\
$18\rightarrow19$ & 0.0340 & 0.0299 & 13.3683 & 926 & 4684 \\
$18\rightarrow20$ & 0.0324 & 0.0291 & 12.5927 & 1178 & 5872 \\
$19\rightarrow13$ & 0.0176 & 0.0152 & 16.5093 & 1000 & 4836 \\
$19\rightarrow14$ & 0.0195 & 0.0166 & 8.7549 & 436 & 1978 \\
$19\rightarrow15$ & 0.0201 & 0.0171 & 15.9232 & 514 & 2562 \\
$19\rightarrow18$ & 0.0275 & 0.0236 & 11.4827 & 393 & 1626 \\
$19\rightarrow20$ & 0.0163 & 0.0139 & 17.1234 & 986 & 5832 \\
$20\rightarrow13$ & 0.0158 & 0.0134 & 17.8276 & 983 & 4804 \\
$20\rightarrow14$ & 0.0249 & 0.0208 & 14.5898 & 450 & 2005 \\
$20\rightarrow15$ & 0.0161 & 0.0134 & 15.9546 & 580 & 2586 \\
$20\rightarrow18$ & 0.0265 & 0.0222 & 13.8473 & 408 & 1625 \\
$20\rightarrow19$ & 0.0154 & 0.0127 & 16.5008 & 839 & 4659 \\
\hline
\end{tabular}}
\caption{\label{tab:context_drift} Summary Statistics of \texttt{Llama-FT} Embeddings for Diachronic Semantic Change of 261 Consistent Lemmas.}
\end{table*}

The aggregate statistics for diachronic semantic drift, computed using \texttt{Llama-FT} contextual embeddings across 261 consistent lemmas from the 13th, 14th, 15th, 18th, 19th, and 20th centuries, are presented in Table \ref{tab:context_drift}. Overall, the results indicate that semantic drift is relatively limited in magnitude. The average global centroid distances range from approximately 0.0154 to 0.0340, and the corresponding average persistent shifts remain consistently close to the original global distances across all time periods analysed. The percentages of drift reduction mostly fall within the range of 11\% to 18\%, with two notable exceptions: $15 \rightarrow 14$ (6.98\%) and $19 \rightarrow 14$ (8.75\%), falling well below this range. This suggests that, with those two exceptions, the majority of semantic representations maintain substantial continuity over the centuries, and that observed semantic changes are primarily influenced by a smaller subset of high-impact contextual usages rather than widespread semantic restructuring.

The most significant semantic divergence is observed in the permutation $18 \rightarrow 19$, which shows the highest average global distance (0.0340) and a persistent shift (0.0299). This indicates that the linguistic contexts represented between these periods have a greater level of contextual and stylistic divergence. More generally, permutations involving the 18th century occupy nearly the entire upper range of the table: all ten permutations pairing the 18th century with another period show global distances at or above 0.0259, well above the cross-permutation mean of 0.0238. This points to the 18th-century sub-corpus as a persistent outlier in contextual and stylistic composition relative to the other five periods.

In contrast, the lowest centroid distances cluster around permutations pairing the 20th century with the 13th, 15th, and 19th centuries: $20\rightarrow19$ (0.0154), $20\rightarrow13$ (0.0158), $20\rightarrow15$ (0.0161), and $19\rightarrow20$ (0.0163) are the four lowest distances. These lower distances indicate stronger semantic continuity linking the 20th century to several earlier and adjacent periods.

The highest average drift reduction percentage occurs in the permutation $20\rightarrow13$ at 17.83\%, closely followed by $13\rightarrow20$ at 17.72\%. These significant reductions indicate that comparisons involving the 13th century are particularly sensitive to isolated high-impact contextual usages, where a relatively small number of innovative, stylistically distinct, or domain-specific cases can disproportionately affect the overall semantic direction of the lemma representations.

The analysis of drifted instance counts uncovers clear trends at the corpus level. Permutations targeting the 20th century collectively account for the largest instance pools (29,299 total instances across the five permutations) and the highest drifted counts (5,564 instances), with four of the five; $13\rightarrow20$, $14\rightarrow20$, $15\rightarrow20$, and $18\rightarrow20$ individually exceeding one thousand drifted instances. Permutations targeting the 19th century show a comparable but somewhat smaller pattern (23,363 total instances, 4,473 drifted). Despite these high absolute counts, several of the corresponding average distances remain among the lowest in the table (e.g. $20\rightarrow19$, $20\rightarrow13$, $20\rightarrow15$), reinforcing the conclusion that while semantic variation becomes more prevalent in modern corpora due to greater textual coverage and lexical diversity, the extent of semantic deviation per instance remains relatively moderate.


The findings indicate that diachronic semantic change in Sinhala, as represented by \texttt{Llama-FT} embeddings, is characterised more by localised contextual innovations than by uniform lexical replacement, with the 18th-century sub-corpus standing out as the most consistent source of elevated divergence across the table. The relatively small differences between global distances and persistent shifts across most permutations suggest that core semantic structures remain largely stable over time, while a minority of significant contextual outliers, concentrated particularly in 13th-century and 18th-century comparisons, account for a disproportionate share of the observed semantic drift.

The semantic trajectory of the lemma "\raisebox{-0.5ex}{%
\includegraphics[height=1.5\fontcharht\font`\A]{Figures/si_132.pdf}}" (meaning "path") shows a distinct divergence between its usage in the 20th century and in the 13th century. In the 20th century, the term already carried a Buddhist doctrinal sense but was situated within a diverse and worldly context. Its nearest associated lemmas include "kingship" (\raisebox{-0.5ex}{%
\includegraphics[height=1.4\fontcharht\font`\A]{Figures/si_188.pdf}}), "long/extended" (\raisebox{-0.5ex}{%
\includegraphics[height=1.5\fontcharht\font`\A]{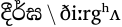}}), "name" (\raisebox{-0.5ex}{%
\includegraphics[height=1.4\fontcharht\font`\A]{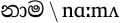}}), "assembly/company" (\raisebox{-0.5ex}{%
\includegraphics[height=1.4\fontcharht\font`\A]{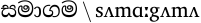}}), and "intercalary-month reckoning" \raisebox{-0.5ex}{%
\includegraphics[height=1.4\fontcharht\font`\A]{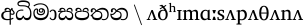}}). This reflects a blend of royal narrative, general description, and calendrical-astrological content. In contrast, the 13th-century context reveals a much narrower focus for "\raisebox{-0.5ex}{%
\includegraphics[height=1.5\fontcharht\font`\A]{Figures/si_132.pdf}}," aligning closely with core meditative and doctrinal vocabulary. This includes terms like "concentration/meditative absorption" (\raisebox{-0.5ex}{%
\includegraphics[height=1.5\fontcharht\font`\A]{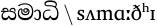}}), "the three planes of existence" (\raisebox{-0.5ex}{%
\includegraphics[height=1.5\fontcharht\font`\A]{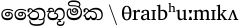}}), "personhood" (\raisebox{-0.5ex}{%
\includegraphics[height=1.5\fontcharht\font`\A]{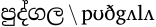}}), and "duty/what ought to be done" (\raisebox{-0.5ex}{%
\includegraphics[height=1.5\fontcharht\font`\A]{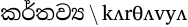}}). The sentences from this period are heavily dominated by a specific canonical work, "\raisebox{-0.5ex}{%
\includegraphics[height=1.5\fontcharht\font`\A]{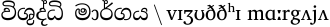}}"\hspace{-1ex} (meaning "the path of purification"). This shift is further supported by the corpus itself, which contains five separate volumes of the "\raisebox{-0.5ex}{%
\includegraphics[height=1.5\fontcharht\font`\A]{Figures/si_217.pdf}}" that date back to the 13th century. The significant factor responsible for the 46.27\% reduction in structural drift is this dense, self-referential commentary context. When comparing the 20th-century context (kingship, description, calendrical themes) with the 13th-century context (concentration, cosmology, personhood), the disparity becomes clear: "\raisebox{-0.5ex}{%
\includegraphics[height=1.5\fontcharht\font`\A]{Figures/si_132.pdf}}" evolves chronologically (13th-20th) from a text-bound reference closely tied to one meditative-commentarial work to a broad doctrinal term embedded in varied worldly discourse.

The semantic trajectory of the lemma "\raisebox{-0.5ex}{%
\includegraphics[height=1.5\fontcharht\font`\A]{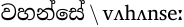}}" (an honorific used for venerable persons, including the Buddha) indicates that in the 13th century, it was situated within a vocabulary related to monastic and genealogical contexts. Words in its vicinity include "\raisebox{-0.5ex}{%
\includegraphics[height=1.5\fontcharht\font`\A]{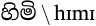}}" (venerable), "\raisebox{-0.5ex}{%
\includegraphics[height=1.5\fontcharht\font`\A]{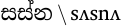}}" (the Buddha's dispensation), "\raisebox{-0.5ex}{%
\includegraphics[height=1.3\fontcharht\font`\A]{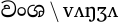}}" (lineage), "\raisebox{-0.5ex}{%
\includegraphics[height=1.3\fontcharht\font`\A]{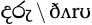}}" (child), and "\raisebox{-0.5ex}{%
\includegraphics[height=1.5\fontcharht\font`\A]{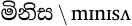}}" (human). This aligns with a blend of devotional and narrative usage that describes the Bodhisattva's quest for enlightenment over innumerable cosmic cycles, as well as ordinary honorific references to respected or deceased individuals. In the 20th-century data, the surrounding vocabulary shifts to include "\raisebox{-0.5ex}{%
\includegraphics[height=1.5\fontcharht\font`\A]{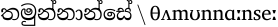}}" (a modern formal honorific pronoun used in correspondence), along with "\raisebox{-0.5ex}{%
\includegraphics[height=1.3\fontcharht\font`\A]{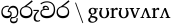}}" (teacher), "\raisebox{-0.5ex}{%
\includegraphics[height=1.4\fontcharht\font`\A]{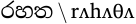}}" (arahant), and "\raisebox{-0.5ex}{%
\includegraphics[height=1.4\fontcharht\font`\A]{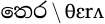}}" (monk). This shift signals a move towards a modern institutional and epistolary register, as evidenced by formal correspondence discussing Christian missionary activity, historical discourse concerning relic enshrinement, and the Buddhist councils, as well as continued use of elaborate devotional names for the Buddha. The primary reason for the 42.67\% reduction in drift is the presence of these dense title chains. The contrast between the 13th-century vocabulary, centred on monastic, lineage, and dispensation terms, and the modern vocabulary that includes honorific pronouns and institutional hierarchy indicates that while "\raisebox{-0.5ex}{%
\includegraphics[height=1.5\fontcharht\font`\A]{Figures/si_218.pdf}}" retains its core honorific function, the surrounding register has visibly modernised and diversified.

\end{document}

%% file: tables/DS.tex
\newcommand{\LitCount}{81}
\newcommand{\DSTable}{ENGALL~\citep{davies2012expanding} & English & 5 & \phantom{-}1800~--~\phantom{-}1999 & $0.09\times10^{13}$ & \tick & \tick & \cross & \tick & \citet{hamilton-etal-2016-cultural} \\
\hline
ENGFIC~\citep{davies2012expanding} & English & 5 & \phantom{-}1800~--~\phantom{-}1999 & $0.07\times10^{12}$ & \tick & \tick & \cross & \tick & \citet{hamilton-etal-2016-cultural} \\
\hline
People in the News~\citep{hennig-wilson-2020-diachronic} & English & 5 & \phantom{-}2000~--~\phantom{-}2019 & $0.02\times10^{11}$ & \tick & \cross & \tick & \tick & \citet{hennig-wilson-2020-diachronic} \\
\hline
COHA~\citep{davies2012expanding} & English & 5 & \phantom{-}1810~--~\phantom{-}2009 & $0.04\times10^{10}$ & \tick & \tick & \tick & \tick & \makecell{\citet{hamilton-etal-2016-diachronic} \\ \citet{giulianelli-etal-2020-analysing} \\ \citet{fonteyn2020grammar} \\ \citet{davies2012expanding}} \\
\hline
EDGeS-English~\citep{bouma-etal-2020-edges} & English & 5 & \phantom{-}1301~--~\phantom{-}2020 & $0.03\times10^{10}$ & \cross & \cross & \cross & \cross &  \\
\hline
English Scientific Writing~\citep{steuer-etal-2024-modeling} & English & 5 & \phantom{-}1665~--~\phantom{-}1996 & $0.03\times10^{10}$ & \tick & \tick & \tick & \cross & \citet{steuer-etal-2024-modeling} \\
\hline
Royal Society Corpus (RSC)~\citep{kermes-etal-2016-royal} & English & 5 & \phantom{-}1665~--~\phantom{-}1869 & $0.04\times10^{9\phantom{0}}$ & \tick & \tick & \tick & \cross & \citet{degaetano-ortlieb-teich-2016-information} \\
\hline
ARCHER~\citep{biber1994archer} & English & 5 & \phantom{-}1700~--~\phantom{-}1994 & $0.03\times10^{8\phantom{0}}$ & \cross & \cross & \cross & \cross &  \\
\hline
PPCHE-Modern British English~\citep{kroch2016penn} & English & 5 & \phantom{-}1707~--~\phantom{-}1914 & $0.03\times10^{8\phantom{0}}$ & \tick & \tick & \tick & \cross & \makecell{\citet{fonteyn2016usage} \\ \citet{fonteyn2025using}} \\
\hline
PPCHE- Early Modern English~\citep{kroch2004penn} & English & 5 & \phantom{-}1500~--~\phantom{-}1720 & $0.02\times10^{8\phantom{0}}$ & \tick & \tick & \tick & \cross & \makecell{\citet{fonteyn2016usage} \\ \citet{fonteyn2025using}} \\
\hline
PPCHE-Middle English~\citep{kroch2000penn} & English & 5 & \phantom{-}1150~--~\phantom{-}1500 & $0.01\times10^{8\phantom{0}}$ & \tick & \tick & \tick & \cross & \makecell{\citet{rodriguez2012development} \\ \citet{tissari2008happiness}} \\
\hline
ISWOC-English~\citep{bech2014iswoc} & English & 5 & \phantom{-}\phantom{0}400~--~\phantom{-}1010 & $0.03\times10^{6\phantom{0}}$ & \cross & \cross & \cross & \cross &  \\
\hline
GERALL~\citep{schneider1998adding} & German & 5 & \phantom{-}1800~--~\phantom{-}1999 & $0.04\times10^{12}$ & \tick & \tick & \tick & \tick & \makecell{\citet{hellrich-hahn-2017-exploring} \\ \citet{hellrich2016measuring}} \\
\hline
German Court Decisions~\citep{braun-2022-tracking} & German & 5 & \phantom{-}1970~--~\phantom{-}2020 & $0.03\times10^{12}$ & \tick & \tick & \tick & \tick & \citet{braun-2022-tracking} \\
\hline
AMC~\citep{jutta2013linguistic} & German & 5 & \phantom{-}1986~--~\phantom{-}2012 & $0.01\times10^{12}$ & \tick & \tick & \tick & \tick & \makecell{\citet{marakasova-neidhardt-2020-short} \\ \citet{baumann2019dylen}} \\
\hline
DTA~\citep{geyken2011deutsche} & German & 5 & \phantom{-}1600~--~\phantom{-}1899 & $0.02\times10^{10}$ & \tick & \tick & \tick & \tick & \makecell{\citet{schlechtweg-etal-2018-diachronic} \\ \citet{schlechtweg-etal-2020-semeval}} \\
\hline
EDGeS-German~\citep{bouma-etal-2020-edges} & German & 5 & \phantom{-}1301~--~\phantom{-}2020 & $0.09\times10^{9\phantom{0}}$ & \cross & \cross & \cross & \cross &  \\
\hline
ParlAT~\citep{wissik2018parlat} & German & 5 & \phantom{-}1945~--~\phantom{-}2017 & $0.07\times10^{9\phantom{0}}$ & \tick & \cross & \cross & \cross & \citet{yim-etal-2022-visualizing} \\
\hline
RIDGES~\citep{odebrecht2017ridges} & German & 5 & \phantom{-}1478~--~\phantom{-}1870 & $0.03\times10^{8\phantom{0}}$ & \cross & \cross & \cross & \cross &  \\
\hline
ReM~\citep{klein2016handbuch} & German & 5 & \phantom{-}1050~--~\phantom{-}1350 & $0.02\times10^{8\phantom{0}}$ & \tick & \tick & \tick & \cross & \citet{breitbarth2019should} \\
\hline
GerManC~\citep{durrell2012germanc} & German & 5 & \phantom{-}1650~--~\phantom{-}1800 & $0.08\times10^{7\phantom{0}}$ & \tick & \tick & \tick & \cross & \makecell{\citet{whitt2016using} \\ \citet{whitt2018evidentiality} \\ \citet{hartmann2014nominalization}} \\
\hline
FREALL~\citep{sagot-etal-2006-lefff} & French & 5 & \phantom{-}1800~--~\phantom{-}1999 & $0.02\times10^{13}$ & \tick & \tick & \cross & \tick & \citet{hamilton-etal-2016-cultural} \\
\hline
LLODIA-French~\citep{armaselu-etal-2024-llodia} & French & 5 & \phantom{-}1690~--~\phantom{-}1918 & $0.06\times10^{8\phantom{0}}$ & \tick & \tick & \tick & \tick & \citet{armaselu2024multilingual} \\
\hline
SRCMF~\citep{lavrentiev2011syntactic} & French & 5 & \phantom{-}\phantom{0}800~--~\phantom{-}1299 & $0.03\times10^{7\phantom{0}}$ & \cross & \cross & \cross & \cross &  \\
\hline
ISWOC-French~\citep{bech2014iswoc} & French & 5 & \phantom{-}1225~--~\phantom{-}1275 & $0.02\times10^{5\phantom{0}}$ & \cross & \cross & \cross & \cross &  \\
\hline
CHIALL~\citep{xue2005penn} & Chinese & 5 & \phantom{-}1950~--~\phantom{-}1999 & $0.06\times10^{12}$ & \tick & \tick & \tick & \tick & \citet{hamilton-etal-2016-diachronic} \\
\hline
ZhShiftEval~\citep{chen-etal-2022-lexicon} & Chinese & 5 & \phantom{-}1953~--~\phantom{-}2003 & $0.06\times10^{10}$ & \tick & \tick & \tick & \tick & \makecell{\citet{periti-tahmasebi-2024-systematic} \\ \citet{goworek-dubossarsky-2026-rethinking}} \\
\hline
People’s Daily~\citep{he-etal-2014-construction} & Chinese & 5 & \phantom{-}1947~--~\phantom{-}1996 & $0.10\times10^{9\phantom{0}}$ & \tick & \tick & \cross & \tick & \citet{mou2015politicize} \\
\hline
CORDE~\citep{shuger2020corpus} & Spanish & 5 & \phantom{-}1472~--~\phantom{-}1975 & $0.03\times10^{10}$ & \tick & \cross & \tick & \cross & \citet{pascual2023evolucion} \\
\hline
Corpus del Español (CdE)~\citep{davies2002corpus} & Spanish & 5 & \phantom{-}1200~--~\phantom{-}1999 & $0.01\times10^{10}$ & \tick & \tick & \tick & \cross & \makecell{\citet{korfhagen2017social} \\ \citet{healey2017evolution}} \\
\hline
CorDECh~\citep{contreras2009hacia} & Spanish & 5 & \phantom{-}1500~--~\phantom{-}1699 & $0.04\times10^{9\phantom{0}}$ & \tick & \cross & \tick & \cross & \makecell{\citet{garrido2025grammaticalization} \\ \citet{sepulveda2025ha}} \\
\hline
EZLN~\citep{gribomont-2023-diachronic-contextual} & Spanish & 5 & \phantom{-}1952~--~\phantom{-}2023 & $0.03\times10^{9\phantom{0}}$ & \tick & \tick & \tick & \tick & \citet{gribomont2025zapatista} \\
\hline
IAC-Spanish~\citep{sanchez-marco-etal-2010-annotation} & Spanish & 5 & \phantom{-}1100~--~\phantom{-}1599 & $0.02\times10^{9\phantom{0}}$ & \cross & \cross & \cross & \cross &  \\
\hline
IMPACT-es~\citep{sanchez2013open} & Spanish & 5 & \phantom{-}1482~--~\phantom{-}1990 & $0.08\times10^{8\phantom{0}}$ & \tick & \tick & \tick & \tick & \citet{melis-etal-2023-change} \\
\hline
PS Post Scriptum-Spanish~\citep{vaamonde2015ps} & Spanish & 5 & \phantom{-}1500~--~\phantom{-}1800 & $0.08\times10^{7\phantom{0}}$ & \tick & \tick & \tick & \cross & \makecell{\citet{garcia-garcia-salido-2019-method} \\ \citet{simonenkosemantic}} \\
\hline
ISWOC-Spanish~\citep{bech2014iswoc} & Spanish & 5 & \phantom{-}1221~--~\phantom{-}1492 & $0.05\times10^{6\phantom{0}}$ & \cross & \cross & \cross & \cross &  \\
\hline
EDGeS-Dutch~\citep{bouma-etal-2020-edges} & Dutch & 4 & \phantom{-}1301~--~\phantom{-}2020 & $0.03\times10^{9\phantom{0}}$ & \tick & \tick & \cross & \cross & \citet{cousse2024auxiliaries} \\
\hline
DiaCORIS~\citep{onelli-etal-2006-diacoris} & Italian & 4 & \phantom{-}1750~--~\phantom{-}1945 & $0.01\times10^{10}$ & \tick & \tick & \tick & \cross & \citet{viola2017corpus} \\
\hline
GDLI ~\citep{favaro-etal-2022-towards} & Italian & 4 & \phantom{-}1300~--~\phantom{-}1999 & $0.03\times10^{6\phantom{0}}$ & \tick & \cross & \tick & \cross & \citet{musi2016semantic} \\
\hline
Kubhist~\citep{lilljegren2018introduktion} & Swedish & 4 & \phantom{-}1700~--~\phantom{-}1999 & $0.03\times10^{11}$ & \tick & \tick & \tick & \tick & \citet{schlechtweg-etal-2020-semeval} \\
\hline
The Swedish Culturomics Gigaword corpus~\citep{eide2016swedish} & Swedish & 4 & \phantom{-}1950~--~\phantom{-}2015 & $0.01\times10^{11}$ & \tick & \tick & \tick & \tick & \citet{ahrenberg2022analysing} \\
\hline
EDGeS-Swedish~\citep{bouma-etal-2020-edges} & Swedish & 4 & \phantom{-}1301~--~\phantom{-}2020 & $0.08\times10^{8\phantom{0}}$ & \cross & \cross & \cross & \cross &  \\
\hline
FSV~\citep{delsing2002fornsvenska} & Swedish & 4 & \phantom{-}1276~--~\phantom{-}1734 & $0.01\times10^{8\phantom{0}}$ & \tick & \tick & \tick & \cross & \citet{sundquist2018diachronic} \\
\hline
Menota-Swedish~\citep{haugen2008menota} & Swedish & 4 & \phantom{-}1400~--~\phantom{-}1550 & $0.03\times10^{7\phantom{0}}$ & \cross & \cross & \cross & \cross &  \\
\hline
HaCOSSA~\citep{hoder2012annotating} & Swedish & 4 & \phantom{-}1375~--~\phantom{-}1550 & $0.01\times10^{7\phantom{0}}$ & \cross & \cross & \cross & \cross &  \\
\hline
MAþiR~\citep{text2024MApiR} & Swedish & 4 & \phantom{-}1200~--~\phantom{-}1299 & $0.03\times10^{6\phantom{0}}$ & \cross & \cross & \cross & \cross &  \\
\hline
CdP~\citep{davies2009creating} & Portuguese & 4 & \phantom{-}1200~--~\phantom{-}1900 & $0.04\times10^{9\phantom{0}}$ & \tick & \cross & \tick & \cross & \citet{gerhalter2025aspectual} \\
\hline
Colonia~\citep{zampieri2013colonia} & Portuguese & 4 & \phantom{-}1500~--~\phantom{-}1999 & $0.06\times10^{8\phantom{0}}$ & \tick & \cross & \cross & \cross & \citet{amaral2023tracing} \\
\hline
TBCHP~\citep{galves2005change} & Portuguese & 4 & \phantom{-}1500~--~\phantom{-}1899 & $0.03\times10^{8\phantom{0}}$ & \tick & \cross & \tick & \cross & \citet{amaral2012nominal} \\
\hline
PS Post Scriptum-Portuguese~\citep{vaamonde2015ps} & Portuguese & 4 & \phantom{-}1500~--~\phantom{-}1800 & $0.06\times10^{7\phantom{0}}$ & \tick & \tick & \tick & \cross & \citet{garcia-garcia-salido-2019-method} \\
\hline
ISWOC-Portuguese~\citep{bech2014iswoc} & Portuguese & 4 & \phantom{-}1344~--~\phantom{-}1400 & $0.04\times10^{6\phantom{0}}$ & \cross & \cross & \cross & \cross &  \\
\hline
DIAKORP~\citep{kuvcera2015diakorp} & Czech & 4 & \phantom{-}1300~--~\phantom{-}1999 & $0.04\times10^{8\phantom{0}}$ & \cross & \cross & \cross & \cross &  \\
\hline
DIALEKT~\citep{koprivova-etal-2014-mapping} & Czech & 4 & \phantom{-}1960~--~\phantom{-}1989 & $0.02\times10^{7\phantom{0}}$ & \cross & \cross & \cross & \cross &  \\
\hline
Menota-Norwegian~\citep{haugen2008menota} & Norwegian & 4 & \phantom{-}1200~--~\phantom{-}1350 & $0.09\times10^{7\phantom{0}}$ & \tick & \cross & \tick & \cross & \citet{vindenes2017complex} \\
\hline
RuSemShift~\citep{rodina-kutuzov-2020-rusemshift} & Russian & 4 & \phantom{-}1682~--~\phantom{-}2017 & $0.03\times10^{10}$ & \tick & \tick & \tick & \tick & \citet{rodina2020elmo} \\
\hline
RuShiftEval~\citep{kutuzov-pivovarova-2021-three} & Russian & 4 & \phantom{-}1700~--~\phantom{-}2016 & $0.04\times10^{7\phantom{0}}$ & \tick & \tick & \tick & \tick & \citet{lidia2021rushifteval} \\
\hline
TOROT-Russian~\citep{eckhoff2015linguistics} & Russian & 4 & \phantom{-}1400~--~\phantom{-}1699 & $0.06\times10^{6\phantom{0}}$ & \tick & \cross & \cross & \cross & \citet{zanchi2016multiple} \\
\hline
HGDS~\citep{simon2014corpus} & Hungarian & 4 & \phantom{-}\phantom{0}900~--~\phantom{-}1499 & $0.03\times10^{8\phantom{0}}$ & \tick & \tick & \tick & \cross & \citet{egedi2012gradual} \\
\hline
RoDICA~\citep{gifu2016tracing} & Romanian & 4 & \phantom{-}1840~--~\phantom{-}1991 & $0.05\times10^{7\phantom{0}}$ & \tick & \tick & \tick & \tick & \citet{truicua2023semantic} \\
\hline
Menota-Danish~\citep{haugen2008menota} & Danish & 3 & \phantom{-}1300~--~\phantom{-}1300 & $0.02\times10^{6\phantom{0}}$ & \cross & \cross & \cross & \cross &  \\
\hline
LLODIA-Hebrew~\citep{armaselu-etal-2024-llodia} & Hebrew & 3 & \phantom{-}1000~--~\phantom{-}2024 & $0.01\times10^{10}$ & \tick & \cross & \cross & \cross & \citet{armaselu-etal-2024-llodia} \\
\hline
PROIEL-Ancient Greek~\citep{haug2008creating} & Greek & 3 & \phantom{-}\phantom{0}\phantom{0}\phantom{0}0~--~\phantom{-}\phantom{0}999 & $0.03\times10^{7\phantom{0}}$ & \tick & \cross & \tick & \cross & \citet{lavidas2020postclassical} \\
\hline
SLIEKKAS~\citep{gelumbeckaite2012senosios} & Lithuanian & 3 & \phantom{-}1500~--~\phantom{-}1800 & $0.04\times10^{7\phantom{0}}$ & \tick & \tick & \tick & \tick & \citet{armaselu-etal-2024-llodia} \\
\hline
IMP-sl~\citep{erjavec2015imp} & Slovene & 3 & \phantom{-}1584~--~\phantom{-}1918 & $0.03\times10^{7\phantom{0}}$ & \cross & \cross & \cross & \cross &  \\
\hline
PROIEL-Armenian~\citep{haug2008creating} & Armenian & 3 & \phantom{-}\phantom{0}400~--~\phantom{-}\phantom{0}450 & $0.02\times10^{6\phantom{0}}$ & \cross & \cross & \cross & \cross &  \\
\hline
LatinISE~\citep{mcgillivray2013tools} & Latin & 3 & \phantom{0}-186~--~\phantom{-}2000 & $0.01\times10^{9\phantom{0}}$ & \tick & \tick & \tick & \tick & \makecell{\citet{mcgillivray2021lexical} \\ \citet{schlechtweg-etal-2020-semeval}} \\
\hline
PROIEL-Latin~\citep{haug2008creating} & Latin & 3 & \phantom{-}\phantom{0}300~--~\phantom{-}\phantom{0}499 & $0.02\times10^{7\phantom{0}}$ & \cross & \cross & \cross & \cross &  \\
\hline
GNC~\citep{gippert2015structuring} & Georgian & 3 & \phantom{-}\phantom{0}400~--~\phantom{-}2015 & $0.02\times10^{10}$ & \cross & \cross & \cross & \cross &  \\
\hline
IcePaHC~\citep{rognvaldsson-etal-2012-icelandic} & Icelandic & 2 & \phantom{-}1100~--~\phantom{-}2012 & $0.01\times10^{8\phantom{0}}$ & \tick & \tick & \tick & \cross & \citet{egedi2013grammatical} \\
\hline
Menota-Icelandic~\citep{haugen2008menota} & Icelandic & 2 & \phantom{-}1200~--~\phantom{-}1700 & $0.09\times10^{7\phantom{0}}$ & \cross & \cross & \cross & \cross &  \\
\hline
Greinir skáldskapar~\citep{eythorsson2014greinir} & Icelandic & 2 & \phantom{-}\phantom{0}900~--~\phantom{-}1270 & $0.03\times10^{6\phantom{0}}$ & \cross & \cross & \cross & \cross &  \\
\hline
Pre-Standard Irish~\citep{scannell-2022-diachronic} & Irish & 2 & \phantom{-}1600~--~\phantom{-}1936 & $0.04\times10^{5\phantom{0}}$ & \cross & \cross & \cross & \cross &  \\
\hline
SiDiaC-v.2.0~\citep{jayatilleke2026sidiacv20sinhaladiachroniccorpus} & Sinhala & 2 & \phantom{-}\phantom{0}400~--~\phantom{-}1999 & $0.02\times10^{7\phantom{0}}$ & \tick & \tick & \cross & \cross & \citet{jayatilleke2026sidiacv20sinhaladiachroniccorpus} \\
\hline
SiDiaC-v.1.0~\citep{jayatilleke2025sidiac} & Sinhala & 2 & \phantom{-}\phantom{0}400~--~\phantom{-}1999 & $0.06\times10^{6\phantom{0}}$ & \cross & \cross & \cross & \cross &  \\
\hline
DCS~\citep{hellwig-etal-2020-treebank} & Sanskrit & 2 & -1300~--~\phantom{0}-700 & $0.05\times10^{7\phantom{0}}$ & \cross & \cross & \cross & \cross &  \\
\hline
FarPaHC~\citep{rognvaldsson-etal-2012-icelandic} & Faroese & 2 & \phantom{-}1800~--~\phantom{-}2012 & $0.05\times10^{6\phantom{0}}$ & \cross & \cross & \cross & \cross &  \\
\hline
DACON~\citep{o2024diachronic} & Newar & 1 & \phantom{-}1114~--~\phantom{-}1899 & $0.02\times10^{6\phantom{0}}$ & \cross & \cross & \cross & \cross &  \\
\hline
DIACU-Old Church Slavonic~\citep{cassese-etal-2025-diacu} & Slavonic & 1 & \phantom{-}\phantom{0}800~--~\phantom{-}1799 & $0.02\times10^{8\phantom{0}}$ & \cross & \cross & \cross & \cross &  \\
\hline
TOROT-Old Church Slavonic~\citep{eckhoff2015linguistics} & Slavonic & 1 & \phantom{-}\phantom{0}800~--~\phantom{-}1099 & $0.02\times10^{7\phantom{0}}$ & \cross & \cross & \cross & \cross &  \\
\hline
PROIEL-Old Church Slavonic~\citep{haug2008creating} & Slavonic & 1 & \phantom{-}\phantom{0}800~--~\phantom{-}1099 & $0.01\times10^{7\phantom{0}}$ & \tick & \cross & \cross & \cross & \citet{eckhoff2014grammatical} \\
\hline
TOROT-Kiev-era Old East Slavic~\citep{eckhoff2015linguistics} & Slavonic & 1 & \phantom{-}\phantom{0}800~--~\phantom{-}1250 & $0.09\times10^{6\phantom{0}}$ & \cross & \cross & \cross & \cross &  \\\hline}